\documentclass[11pt,a4paper]{article}
\usepackage[hyperref]{acl2020}
\usepackage{times}
\usepackage{latexsym}

\usepackage{microtype}

\aclfinalcopy 
\def\aclpaperid{2161} 

\usepackage{graphicx} 
\usepackage{float} 
\usepackage{subfigure} 
\usepackage{amsmath}
\usepackage{color}
\usepackage{array}
\usepackage{xspace}
\usepackage{url}
\usepackage{wrapfig}
\usepackage{algorithm,algorithmic}
\usepackage{trackchanges}
\addeditor{YJ}
\newfloat{algorithm}{t}{lop}

\usepackage{amsmath,amsfonts,amssymb,bm}
\usepackage{pifont} 
\usepackage{graphicx,subfigure,epsfig,fancybox} 
\usepackage{float}
\usepackage{color} 
\usepackage{multirow}
\usepackage{natbib}

\usepackage{tikz}
\usetikzlibrary{shapes.geometric, positioning, calc}
\usetikzlibrary{arrows,shapes,calc}
\usetikzlibrary{trees,positioning,mindmap,shadows,fit}
\usetikzlibrary{decorations.pathreplacing}
\usetikzlibrary{tikzmark} 
\usetikzlibrary{intersections} 

\newcommand{\revised}[1]{\textcolor{blue}{}}
\renewcommand{\vec}[1]{\boldsymbol{#1}}

\usepackage{adjustbox}
\usepackage{array}
\usepackage{booktabs}

\newcolumntype{R}[2]{%
    >{\adjustbox{angle=#1,lap=\width-(#2)}\bgroup}%
    l%
    <{\egroup}%
}

\usepackage{todonotes}
\usepackage{setspace}

\newcommand{\ourmethod}{\textsc{Hedge}\xspace}
\newcommand{\tmptitle}{Generating Hierarchical Explanations on Text Classification via Feature Interaction Detection}

\title{\tmptitle}

\author{}

\author{Hanjie Chen,  Guangtao Zheng, Yangfeng Ji\\
	Department of Computer Science \\
	University of Virginia \\
	Charlottesville, VA, USA \\
	\texttt{\{hc9mx, gz5hp, yangfeng\}@virginia.edu} \\}

\date{}

\begin{document}
\maketitle

\begin{abstract}
  Generating explanations for neural networks has become crucial for their applications in real-world with respect to reliability and trustworthiness. In natural language processing, existing methods usually provide important features which are words or phrases selected from an input text as an explanation, but ignore the interactions between them. 
It poses challenges for humans to interpret an explanation and connect it to model prediction. 
In this work, we build hierarchical explanations by detecting feature interactions. 
Such explanations visualize how words and phrases are combined at different levels of the hierarchy, which can help users understand the decision-making of black-box models. 
The proposed method is evaluated with three neural text classifiers (LSTM, CNN, and BERT) on two benchmark datasets, via both automatic and human evaluations. 
Experiments show the effectiveness of the proposed method in providing explanations that are both faithful to models and interpretable to humans.
\end{abstract}

\section{Introduction}
\label{sec:intro}
Deep neural networks have achieved remarkable performance in natural language processing (NLP)~\citep{devlin2018bert, howard2018universal, peters2018deep}, 
but the lack of understanding on their decision making leads them to be characterized as \emph{blackbox models} and increases the risk of applying them in real-world applications~\citep{lipton2016mythos, burns2018exploiting, jumelet2018language, jacovi2018understanding}.

Understanding model prediction behaviors has been a critical factor in whether people will trust and use these blackbox models~\citep{ribeiro2016should}. 
A typical work on understanding decision-making is to generate prediction explanations for each input example, called local explanation generation. In NLP, most of existing work on local explanation generation focuses on producing word-level or phrase-level explanations by quantifying contributions of individual words or phrases to a model prediction~\citep{ribeiro2016should,lundberg2017unified,lei2016rationalizing,plumb2018model}.
\begin{figure}[tbph]
  \centering
  \includegraphics[width=0.45\textwidth]{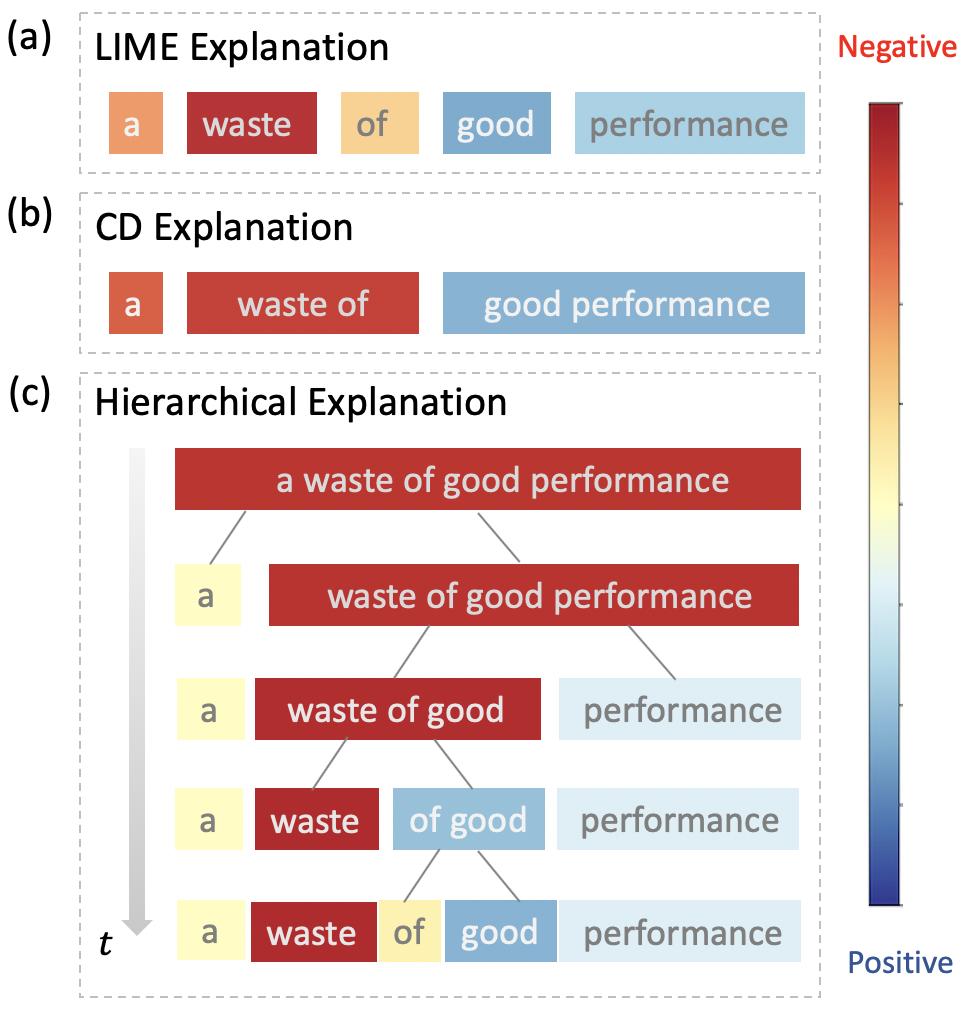}
  \setlength{\belowcaptionskip}{-10pt}
  \caption{\label{fig:explain_exps} Different explanations for a \textsc{negative} movie review \texttt{a waste of good performance}, where the color of each block represents the contribution of the corresponding word/phrase/clause (feature) to the model prediction. From the hierarchical explanation, we obtain a set of features in each timestep ($t$), where the most important one is \texttt{waste of good}.}
\end{figure}

\autoref{fig:explain_exps} (a) and (b) present a word-level and a phrase-level explanation generated by the LIME \citep{ribeiro2016should} and the Contextual Decomposition (CD)~\citep{murdoch2018beyond} respectively for explaining sentiment classification.
Both explanations provide scores to quantify how a word or a phrase contributes to the final prediction.
For example, the explanation generated by LIME captures a keyword \texttt{waste} and the explanation from CD identifies an important phrase \texttt{waste of}.
However, neither of them is able to explain the model decision-making in terms of how words and phrases are interacted with each other and composed together for the final prediction. 
In this example, since the final prediction is \textsc{negative}, one question that we could ask is that how the word \texttt{good} or a phrase related to the word \texttt{good} contributes to the model prediction.
An explanation being able to answer this question will give users a better understanding on the model decision-making and also more confidence to trust the prediction.

The goal of this work is to reveal prediction behaviors of a text classifier by detecting feature (e.g., words or phrases) interactions with respect to model predictions.
For a given text, we propose a model-agnostic approach, called \ourmethod (for Hierarchical Explanation via Divisive Generation), to build hierarchical explanations by recursively detecting the weakest interactions and then dividing large text spans into smaller ones based on the interactions.
As shown in \autoref{fig:explain_exps} (c), the hierarchical structure produced by \ourmethod provides a comprehensive picture of how different granularity of features interacting with each other within the model.
For example, it shows how the word \texttt{good} is dominated by others in the model prediction, which eventually leads to the correct prediction.
Furthermore, the scores of text spans across the whole hierarchy also help identify the most important feature \texttt{waste of good}, which can be served as a phrase-level explanation for the model prediction.


The contribution of this work is three-fold:
(1) we design a top-down model-agnostic method of constructing hierarchical explanations via feature interaction detection;
(2) we propose a simple and effective scoring function to quantify feature contributions with respect to model predictions;
and (3) we compare the proposed algorithm with several competitive methods on explanation generation via both automatic and human evaluations.
The experiments were conducted on sentiment classification tasks with three neural network models, LSTM~\cite{hochreiter1997long}, CNN~\cite{kim2014convolutional}, and BERT~\cite{devlin2018bert}, on the SST~\citep{socher2013recursive} and IMDB~\citep{maas2011learning} datasets.
The comparison with other competitive methods illustrates that \ourmethod provides more faithful and human-understandable explanations.

Our implementation is available at \url{https://github.com/UVa-NLP/HEDGE}.

\section{Related Work}
\label{sec:relate}
Over the past years, many approaches have been explored to interpret neural networks, such as contextual decomposition (CD) for LSTM~\citep{murdoch2018beyond} or CNN model~\citep{godin2018explaining}, gradient-based interpretation methods~\citep{hechtlinger2016interpretation,sundararajan2017axiomatic}, and attention-based methods~\citep{ghaeini2018interpreting, lee2017interactive, serrano2019attention}. However, these methods have limited capacity in real-world applications, as they require deep understanding of neural network architectures \citep{murdoch2018beyond} or only work with specific models \citep{alvarez2018towards}. On the other hand, model-agnostic methods \citep{ribeiro2016should,lundberg2017unified} generate explanations solely based on model predictions and are applicable for any black-box models. In this work, we mainly focus on model-agnostic explanations.

\subsection{Model-Agnostic Explanations}
\label{subsec:model_ago}
The core of generating model-agnostic explanations is how to efficiently evaluate the importance of features with respect to the prediction.
So far, most of existing work on model-agnostic explanations focus on the word level.
For example, \citet{li2016understanding} proposed Leave-one-out to probe the black-box model by observing the probability change on the predicted class when erasing a certain word.
LIME proposed by \citet{ribeiro2016should} estimates individual word contribution locally by linear approximation from perturbed examples.
A line of relevant works to ours is Shapley-based methods, where the variants of Shapley values~\citep{shapley1953value} are used to evaluate feature importance, such as SampleShapley~\citep{kononenko2010efficient}, KernelSHAP~\citep{lundberg2017unified}, and L/C-Shapley~\citep{chen2018shapley}.
They are still in the category of generating word-level explanations, while mainly focus on addressing the challenge of computational complexity of Shapley values~\citep{datta2016algorithmic}.
In this work, inspired by an extension of Shapley values~\citep{owen1972multilinear, grabisch1997k, fujimoto2006axiomatic}, we design a function to detect feature interactions for building hierarchical model-agnostic explanations in \autoref{subsec:top-down}.
While, different from prior work of using Shapley values for feature importance evaluation, we propose an effective and simpler way to evaluate feature importance as described in \autoref{subsec:fea-import}, which outperforms Shapley-based methods in selecting important words as explanations in \autoref{subsec:quantieva}.

\subsection{Hierarchical Explanations}
\label{subsec:hier_inter}
Addressing the limitation of word-level explanations (as discussed in \autoref{sec:intro}) has motivated the work on generating phrase-level or hierarchical explanations. 
For example, \citet{tsang2018can} generated hierarchical explanations by considering the interactions between any features with exhaustive search, which is computationally expensive.

\citet{singh2018hierarchical} proposed agglomerative contextual decomposition (ACD) which utilizes CD scores~\citep{murdoch2018beyond, godin2018explaining} for feature importance evaluation and employ a hierarchical clustering algorithm to aggregate features together for hierarchical explanation. Furthermore, \citet{jin2019hierarchical} indicated the limitations of CD and ACD in calculating phrase interactions in a formal context, and proposed two explanation algorithms by quantifying context independent importance of words and phrases. 

A major component of the proposed method on feature interaction detection is based on the Shapley interaction index~\citep{owen1972multilinear, grabisch1997k, fujimoto2006axiomatic}, which is extended in this work to capture the interactions in a hierarchical structure. 
 \citet{lundberg2018consistent} calculated features interactions via SHAP interaction values along a given tree structure. \citet{chen2019ls} suggested to utilize a linguistic tree structure to capture the contributions beyond individual features for text classification.
The difference with our work is that both methods \citep{lundberg2018consistent, chen2019ls} require hierarchical structures given, while our method constructs structures solely based on feature interaction detection without resorting external structural information. 
In addition, different from \citet{singh2018hierarchical}, our algorithm uses a top-down fashion to divide long texts into short phrases and words based on the weakest interactions, which is shown to be more effective and efficient in the experiments in \autoref{sec:exp}.

\section{Method}
\label{sec:method}
This section explains the proposed algorithm on building hierarchical explanations (\autoref{subsec:top-down}) and two critical components of this algorithm: detecting feature interaction (\autoref{subsec:fea-interac}) and quantifying feature importance (\autoref{subsec:fea-import}).

\begin{algorithm}[th!]
  \caption{Hierarchical Explanation via Divisive Generation}
  \label{alg:hedge}
  \begin{algorithmic}[1]
    \STATE {\bf Input}: text $\vec{x}$ with length $n$, and predicted label $\hat{y}$
    \STATE Initialize the original partition $\mathcal{P}_{0} \leftarrow \{\vec{x}_{(0,n]}\}$
    \STATE Initialize the contribution set $\mathcal{C}_0 = \emptyset$
    \STATE Initialize the hierarchy $\mathcal{H}=[\mathcal{P}_0]$
    \FOR {$t = 1,\dots,n-1$}
    \STATE Find $\vec{x}_{(s_i,s_{i+1}]}$ and $j$ by solving \autoref{eq:obj_func}
    \STATE Update the partition\\
    $\mathcal{P}_t'\leftarrow \mathcal{P}_{t-1}\backslash\{\vec{x}_{(s_i,s_{i+1}]}\}$\\
    $\mathcal{P}_t\leftarrow \mathcal{P}'_t \cup \{\vec{x}_{(s_i,j]}, \vec{x}_{(j,s_{i+1}]}\}$
    \STATE $\mathcal{H}.add(\mathcal{P}_t)$
    \STATE Update the contribution set $\mathcal{C}$ with\\
    $\mathcal{C}_t'\leftarrow \mathcal{C}_{t-1}\cup \{(\vec{x}_{(s_i,j]},\psi(\vec{x}_{(s_i,j]}))\}$\\
    $\mathcal{C}_t\leftarrow \mathcal{C}_{t}'\cup \{(\vec{x}_{(j,s_{i+1}]},\psi(\vec{x}_{(j,s_{i+1}]}))\}$
    \ENDFOR
    \STATE {\bf Output}: $\mathcal{C}_{n-1}$, $\mathcal{H}$
  \end{algorithmic}
\end{algorithm}
\subsection{Generating Hierarchical Explanations}
\label{subsec:top-down}
For a classification task, let $\vec{x}=(x_1,\dots,x_n)$ denote a text with $n$ words and $\hat{y}$ be the prediction label from a well-trained model. 
Furthermore, we define $\mathcal{P}=\{\vec{x}_{(0, s_1]},\vec{x}_{(s_1, s_2]},\dots, \vec{x}_{(s_{P-1}, n]}\}$ be a partition of the word sequence with $P$ text spans, where $\vec{x}_{(s_i,s_{i+1}]}=(x_{s_i+1},\dots,x_{s_{i+1}})$.
For a given text span $\vec{x}_{(s_i,s_{i+1}]}$, the basic procedure of \ourmethod is to divide it into two smaller text spans $\vec{x}_{(s_i,j]}$ and $\vec{x}_{(j,s_{i+1}]}$, where $j$ is the dividing point ($s_i< j < s_{i+1}$), and then evaluate their contributions to the model prediction $\hat{y}$.

Algorithm~\ref{alg:hedge} describes the whole procedure of dividing $\vec{x}$ into different levels of text spans and evaluating the contribution of each of them.
Starting from the whole text $\vec{x}$, the algorithm first divides $\vec{x}$ into two segments.
In the next iteration, it will pick one of the two segments and further split it into even smaller spans.
As shown in algorithm~\ref{alg:hedge}, to perform the top-down procedure, we need to answer the questions: for the next timestep, which text span the algorithm should pick to split and where is the dividing point?

Both questions can be addressed via the following optimization problem:
\begin{equation}
  \label{eq:obj_func}
  \min_{\vec{x}_{(s_i,s_{i+1}]}\in\mathcal{P}}\min_{j\in(s_i,s_{i+1})}\phi(\vec{x}_{(s_i,j]}, \vec{x}_{(j,s_{i+1}]}\mid \mathcal{P}),
\end{equation}
where $\phi(\vec{x}_{(s_i,j]}, \vec{x}_{(j,s_{i+1}]}\mid \mathcal{P})$ defines the interaction score between $\vec{x}_{(s_i,j]}$ and $\vec{x}_{(j,s_{i+1}]}$ given the current partition $\mathcal{P}$.
The detail of this score function will be explained in \autoref{subsec:fea-interac}.

For a given $\vec{x}_{(s_i,s_{i+1}]}\in\mathcal{P}$, the inner optimization problem will find the \emph{weakest} interaction point to split the text span $\vec{x}_{(s_i,s_{i+1}]}$ into two smaller ones.
It answers the question about where the dividing point should be for a given text span.
A trivial case of the inner optimization problem is on a text span with length 2, since there is only one possible way to divide it. 
The outer optimization answers the question about which text span should be picked.
This optimization problem can be solved by simply enumerating all the elements in a partition $\mathcal{P}$.
A special case of the outer optimization problem is at the first iteration $t=1$, where $\mathcal{P}_0=\{\vec{x}_{(0,n]}\}$ only has one element, which is the whole input text. 
Once the partition is updated, it is then added to the hierarchy $\mathcal{H}$.

The last step in each iteration is to evaluate the contributions of the new spans and update the contribution set $\mathcal{C}$ as in line 9 of the algorithm~\ref{alg:hedge}.
For each, the algorithm evaluates its contribution to the model prediction with the feature importance function $\psi(\cdot)$ defined in \autoref{eq:local-importance-score}. The final output of algorithm~\ref{alg:hedge} includes the contribution set $\mathcal{C}_{n-1}$ which contains all the produced text spans in each timestep together with their importance scores, and the hierarchy $\mathcal{H}$ which contains all the partitions of $\vec{x}$ along timesteps. A hierarchical explanation can be built based on $\mathcal{C}_{n-1}$ and $\mathcal{H}$ by visualizing the partitions with all text spans and their importance scores along timesteps, as \autoref{fig:explain_exps} (c) shows.

Note that with the feature interaction function $\phi(\cdot,\cdot)$, we could also design a bottom-up approach to merge two short text spans if they have the strongest interaction.
Empirically, we found that this bottom-up approach performs worse than the algorithm~\ref{alg:hedge}, as shown in \autoref{sec:compare-top-bottom}.

\subsection{Detecting Feature Interaction}
\label{subsec:fea-interac}
For a given text span $\vec{x}_{(s_i,s_{i+1}]}\in\mathcal{P}$ and the dividing point $j$, the new partition will be $\mathcal{N}=\mathcal{P}\backslash\{\vec{x}_{(s_i,s_{i+1}]}\}\cup \{\vec{x}_{(s_i,j]}, \vec{x}_{(j,s_{i+1}]}\}=\{\vec{x}_{(0, s_1]},\dots,\vec{x}_{(s_i,j]},\vec{x}_{(j,s_{i+1}]},\dots,\vec{x}_{(s_{P-1}, n]}\}$. We consider the effects of other text spans in $\mathcal{N}$ when calculate the interaction between $\vec{x}_{(s_i,j]}$ and $\vec{x}_{(j,s_{i+1}]}$, since the interaction between two words/phrases is closely dependent on the context \citep{hu2016natural, chen2016enhanced}. We adopt the Shapley interaction index from coalition game theory~\citep{owen1972multilinear, grabisch1997k, fujimoto2006axiomatic} to calculate the interaction. For simplicity, we denote $\vec{x}_{(s_i,j]}$ and $\vec{x}_{(j,s_{i+1}]}$ as $j_1$ and $j_2$ respectively. The interaction score is defined as \citep{lundberg2018consistent},

\begin{small}
\begin{equation}
\label{eq:shapleyinteraction}
\phi(j_1,\! j_2\mid\! \mathcal{P})\! =\sum_{S\subseteq \mathcal{N}\backslash\{j_1, j_2\}}\!\frac{|S|!(P-|S|-1)!}{P!}\gamma(j_1,\! j_2,\! S),
\end{equation}
\end{small}%
where $S$ represents a subset of text spans in $\mathcal{N}\backslash\{j_1,j_2\}$, $|S|$ is the size of $S$, and $\gamma(j_1, j_2, S)$ is defined as follows,

\begin{small}
\begin{equation}
\begin{aligned}
\label{eq:fea_attribution}
\gamma(j_1,\! j_2,\! S) &=\mathbb{E}[f(\vec{x}')\!\mid\! S\cup\! \{j_1,\! j_2\}]-\mathbb{E}[f(\vec{x}')\!\mid\! S\cup\! \{j_1\}]\\
   &-\mathbb{E}[f(\vec{x}')\mid S\cup \{j_2\}]+\mathbb{E}[f(\vec{x}')\mid S],
\end{aligned}
\end{equation}
\end{small}%
where $\vec{x}'$ is the same as $\vec{x}$ except some missing words that are not covered by the given subset (e.g. $S$), $f(\cdot)$ denotes the model output probability on the predicted label $\hat{y}$, and $\mathbb{E}[f(\vec{x}')\mid S]$ is the expectation of $f(\vec{x}')$ over all possible $\vec{x}'$ given $S$. In practice, the missing words are usually replaced with a special token \texttt{<pad>}, and $f(\vec{x}')$ is calculated to estimate $\mathbb{E}[f(\vec{x}')|S]$ \citep{chen2018shapley, datta2016algorithmic, lundberg2017unified}. We also adopt this method in our experiments. Another way to estimate the expectation is to replace the missing words with substitute words randomly drawn from the full dataset, and calculate the empirical mean of all the sampling data~\citep{kononenko2010efficient, vstrumbelj2014explaining}, which has a relatively high computational complexity.

With the number of text spans (features) increasing, the exponential number of model evaluations in \autoref{eq:shapleyinteraction} becomes intractable. We calculate an approximation of the interaction score based on the assumption \citep{chen2018shapley, singh2018hierarchical, jin2019hierarchical}: {a word or phrase usually has strong interactions with its neighbours in a sentence}. The computational complexity can be reduced to polynomial by only considering $m$ neighbour text spans of $j_1$ and $j_2$ in $\mathcal{N}$. The interaction score is rewritten as

\begin{small}
\begin{equation}
\label{eq:shapleyinteraction_1}
\phi(j_1,\! j_2\! \mid\! \mathcal{P})\! =\!\sum_{S\subseteq \mathcal{N}_{m}\backslash\!\{j_1,\! j_2\}}\!\frac{|S|!(M-|S|-2)!}{(M-1)!}\gamma(j_1,\! j_2,\! S),
\end{equation}
\end{small}%
where $\mathcal{N}_{m}$ is the set containing $j_1$, $j_2$ and their neighbours, and $M=|\mathcal{N}_{m}|$. In \autoref{sec:exp}, we set $m=2$, which performs well. The performance can be further improved by increasing $m$, but at the cost of increased computational complexity.

\subsection{Quantifying Feature Importance}
\label{subsec:fea-import}
To measure the contribution of a feature $\vec{x}_{(s_i,s_{i+1}]}$ to the model prediction, we define the importance score as

\begin{equation}
\label{eq:local-importance-score}
\begin{split}
    \psi(\vec{x}_{(s_i,s_{i+1}]}) = & f_{\hat{y}}(\vec{x}_{(s_i,s_{i+1}]})\\
     & -\max_{y'\neq \hat{y}, y'\in\mathcal{Y}}f_{y'}(\vec{x}_{(s_i,s_{i+1}]}),
\end{split}
\end{equation}
where $f_{\hat{y}}(\vec{x}_{(s_i,s_{i+1}]})$ is the model output on the predicted label $\hat{y}$; $\max_{y'\neq \hat{y}, y'\in\mathcal{Y}}f_{y'}(\vec{x}_{(s_i,s_{i+1}]})$ is the highest model output among all classes excluding $\hat{y}$. This importance score measures how far the prediction on a given feature is to the prediction boundary, hence the confidence of classifying $\vec{x}_{(s_i,s_{i+1}]}$ into the predicted label $\hat{y}$. 
Particularly in text classification, it can be interpreted as the contribution to a specific class $\hat{y}$.
The effectiveness of \autoref{eq:local-importance-score} as feature importance score is verified in \autoref{subsec:quantieva}, where \ourmethod outperforms several competitive baseline methods (e.g. LIME~\citep{ribeiro2016should}, SampleShapley~\citep{kononenko2010efficient}) in identifying important features.

\section{Experiments}
\label{sec:exp}
The proposed method is evaluated on text classification tasks with three typical neural network models, a long short-term memories~\citep[LSTM]{hochreiter1997long}, a convolutional neural network~\citep[CNN]{kim2014convolutional}, and BERT~\citep{devlin2018bert}, on the SST~\citep{socher2013recursive} and IMDB~\citep{maas2011learning} datasets, via both automatic and human evaluations.

\subsection{Setup}
\label{subsec:setup}
\paragraph{Datasets.} We adopt the SST-2 \citep{socher2013recursive} which has 6920/872/1821 examples in the \textit{train/dev/test} sets with binary labels. The IMDB \citep{maas2011learning} also has binary labels with 25000/25000 examples in the \textit{train/test} sets. We hold out 10\% of the training examples as the development set.
\paragraph{Models.}
The CNN model~\cite{kim2014convolutional} includes a single convolutional layer with filter sizes ranging from 3 to 5. The LSTM~\cite{hochreiter1997long} has a single layer with 300 hidden states. Both models are initialized with 300-dimensional pretrained word embeddings~\cite{mikolov2013distributed}.
We use the pretrained BERT model\footnote{\url{https://github.com/huggingface/pytorch-transformers}{}} with 12 transformer layers, 12 self-attention heads, and the hidden size of 768, which was then fine-tuned with different downstream tasks to achieve the best performance.
Table~\ref{tab:models} shows the best performance of the models on both datasets in our experiments, where BERT outperforms CNN and LSTM with higher classification accuracy.
 \begin{table}[tbph]
  \centering
  \setlength{\belowcaptionskip}{-15pt}
  \begin{tabular}{lll}
     \toprule
 \multirow{2}{*}{Models}    & \multicolumn{2}{c}{Dataset}\\
     \cmidrule{2-3}
      & SST & IMDB \\
     \midrule
     LSTM & 0.842 & 0.870 \\
     CNN & 0.850  & 0.901  \\
     BERT & 0.924 & 0.930 \\
     \bottomrule
  \end{tabular}
  \caption{The classification accuracy of different models on the SST and IMDB datasets.}
  \label{tab:models}
\end{table}



\begin{table*}[tbph]
	\centering
	\begin{tabular}{p{0.1\textwidth}p{0.15\textwidth}cccccc}
		\toprule
	\multirow{2}{*}{Datasets} 	& \multirow{2}{*}{Methods}& \multicolumn{2}{c}{LSTM} & \multicolumn{2}{c}{CNN} & \multicolumn{2}{c}{BERT} \\
		\cmidrule(lr){3-4} \cmidrule(lr){5-6} \cmidrule(lr){7-8}
		&  & AOPC & Log-odds & AOPC & Log-odds & AOPC & Log-odds \\
		\midrule
\multirow{8}{*}{SST} & Leave-one-out & 0.441  & -0.443  & 0.434  & -0.448  & 0.464  & -0.723 \\
		& CD & 0.384  & -0.382  & -  & -  &  - & - \\
		& LIME & 0.444  & -0.449  & 0.473  & -0.542  & 0.134  & -0.186 \\
		& L-Shapley & 0.431  & -0.436  & 0.425  & -0.459  & 0.435  & -0.809 \\
		& C-Shapley & 0.423  & -0.425  & 0.415  &  -0.446 & 0.410  & -0.754 \\
		& KernelSHAP & 0.360  & -0.361  & 0.387  & -0.423  & 0.411  & -0.765 \\
		& SampleShapley & 0.450  & -0.454  & 0.487  & -0.550  & 0.462  & -0.836 \\
		& \ourmethod & \textbf{0.458}  & \textbf{-0.466}  & \textbf{0.494}  & \textbf{-0.567}  & \textbf{0.479}  & \textbf{-0.862} \\
		\midrule
		\multirow{8}{*}{IMDB} & Leave-one-out & 0.630  & -1.409  & 0.598  & -0.806  & 0.335  & -0.849 \\
		& CD & 0.495  & -1.190  & -  & -  & -  & - \\
		& LIME & 0.764  & -1.810  & 0.691  & -1.091  & 0.060  & -0.133 \\
		& L-Shapley & 0.637  & -1.463  & 0.623  & -0.950  & 0.347  & -1.024 \\
		& C-Shapley & 0.629  & -1.427  & 0.613  & -0.928  &  0.331 & -0.973 \\
		& KernelSHAP & 0.542  & -1.261  & 0.464  & -0.727  & 0.223  & -0.917 \\
		& SampleShapley & 0.757  & -1.597  & 0.707  & -1.108  & 0.355  & -1.037 \\
		& \ourmethod & \textbf{0.783}  & \textbf{-1.873}  & \textbf{0.719}  &  \textbf{-1.144} & \textbf{0.411}  & \textbf{-1.126} \\
		\bottomrule
	\end{tabular}
	\caption{AOPCs and log-odds scores of different interpretation methods in explaining different models on the SST and IMDB datasets.}
	\label{tab:aopc-log}
\end{table*}

\subsection{Quantitative Evaluation}
\label{subsec:quantieva}
We adopt two metrics from prior work on evaluating word-level explanations: the area over the perturbation curve (AOPC)~\citep{nguyen2018comparing, samek2016evaluating} and the log-odds scores~\citep{shrikumar2017learning, chen2018shapley}, and define a new evaluation metric called \textit{cohesion-score} to evaluate the interactions between words within a given text span. The first two metrics measure local fidelity by deleting or masking top-scored words and comparing the probability change on the predicted label. They are used to evaluate \autoref{eq:local-importance-score} in quantifying feature contributions to the model prediction. The cohesion-score measures the synergy of words within a text span to the model prediction by shuffling the words to see the probability change on the predicted label. 

\paragraph{AOPC.}
 By deleting top $k\%$ words, AOPC calculates the average change in the prediction probability on the predicted class over all test data as follows,
 
\begin{small}
\begin{equation}
  \label{eq:aopc}
   \setlength{\abovedisplayskip}{-6pt}
   \setlength{\belowdisplayskip}{-0.1pt}
  \text{AOPC}(k)=\frac{1}{N}\sum_{i=1}^N\{p(\hat{y}\mid\vec{x}_{i})-p(\hat{y}\mid \tilde{\vec{x}}^{(k)}_{i})\},
\end{equation}
\end{small}%
where $\hat{y}$ is the predicted label, $N$ is the number of examples, $p(\hat{y}\mid\cdot)$ is the probability on the predicted class, and $\tilde{\vec{x}}^{(k)}_{i}$ is constructed by dropping the $k\%$ top-scored words from $\vec{x}_i$. Higher AOPCs are better, which means that the deleted words are important for model prediction. To compare with other word-level explanation generation methods under this metric, we select word-level features from the bottom level of a hierarchical explanation and sort them in the
order of their estimated importance to the prediction.

\paragraph{Log-odds.}
Log-odds score is calculated by averaging the difference of negative logarithmic probabilities on the predicted class over all of the test data before and after masking the top $r\%$ features with zero paddings,

\begin{small}
\begin{equation}
  \label{eq:logodds}
  \setlength{\abovedisplayskip}{-6pt}
  \setlength{\belowdisplayskip}{-0.1pt}
  \text{Log-odds}(r)
  =\frac{1}{N}\sum_{i=1}^N\log\frac{p(\hat{y}\mid\tilde{\vec{x}}^{(r)}_{i})}{p(\hat{y}\mid\vec{x}_{i})}.
\end{equation}
\end{small}%
The notations are the same as in \autoref{eq:aopc} with the only difference that  $\tilde{\vec{x}}^{(r)}_{i}$ is constructed by replacing the top $r\%$ word features with the special token $\langle \texttt{pad}\rangle$ in $\vec{x}_{i}$.
Under this metric, lower log-odds scores are better.

\paragraph{Cohesion-score.}
We propose cohesion-score to justify an important text span identified by \ourmethod.
Given an important text span $\vec{x}_{(a,b]}$, we randomly pick a position in the word sequence $(x_1,\ldots,x_a,x_{b+1},\ldots,x_n)$ and insert a word back. The process is repeated until a shuffled version of the original sentence $\bar{\vec{x}}$ is constructed.
The cohesion-score is the difference between $p(\hat{y}\mid \vec{x})$ and $p(\hat{y}\mid \bar{\vec{x}})$.
Intuitively, the words in an important text span have strong interactions. By perturbing such interactions, we expect to observe the output probability decreasing.
To obtain a robust evaluation, for each sentence $\vec{x}_i$, we construct $Q$ different word sequences $\{\bar{\vec{x}}_{i}^{(q)}\}_{q=1}^{Q}$ and compute the average as

\begin{small}
\begin{equation}
\label{eq:cohe}
\setlength{\abovedisplayskip}{-6pt}
\setlength{\belowdisplayskip}{-0.1pt}
\text{Cohesion-score}=\frac{1}{N}\sum_{i=1}^N\frac{1}{Q}\sum_{q=1}^Q(p(\hat{y}\mid\vec{x}_{i})-p(\hat{y}\mid\bar{\vec{x}}_{i}^{(q)})),
\end{equation}
\end{small}%
where $\bar{\vec{x}}_{i}^{(q)}$ is the $q^{\text{th}}$ perturbed version of $\vec{x}_{i}$, $Q$ is set as 100, and the most important text span in the contribution set $\mathcal{C}$ is considered. Higher cohesion-scores are better.  

\subsubsection{Results}
We compare \ourmethod with several competitive baselines, namely Leave-one-out~\citep{li2016understanding}, LIME~\citep{ribeiro2016should}, CD~\citep{murdoch2018beyond}, Shapley-based methods, \citep[L/C-Shapley]{chen2018shapley}, \citep[KernelSHAP]{lundberg2017unified}, and \citep[SampleShapley]{kononenko2010efficient}, using AOPC and log-odds metrics; and use cohesion-score to compare \ourmethod with another hierarchical explanation generation method ACD~\citep{singh2018hierarchical}. 

 \begin{table}[tbph]
	\centering
	\begin{tabular}{llll}
		\toprule
		\multirow{2}{*}{Methods} & \multirow{2}{*}{Models} & \multicolumn{2}{c}{Cohesion-score} \\
		\cline{3-4}
		& & SST & IMDB \\
		\midrule
\multirow{3}{*}{\ourmethod}	& CNN & 0.016 & 0.012 \\ 
	& BERT & 0.124 & 0.103 \\
		& LSTM &0.020 & 0.050 \\[0.2cm]
        ACD&  LSTM & 0.015 & 0.038\\
		\bottomrule
	\end{tabular}
	\caption{Cohesion scores of \ourmethod and ACD in interpreting different models on the SST and IMDB datasets. For ACD, we adopt the existing application from the original paper \citep{singh2018hierarchical} to explain LSTM on text classification.}
	\label{tab:phrase-level-evaluation}
\end{table}

The AOPCs and log-odds scores on different models and datasets are shown in \autoref{tab:aopc-log}, where $k=r=20$. 
Additional results of AOPCs and log-odds changing with different $k$ and $r$ are shown in \autoref{sec:other_aopc_log}. For the IMDB dataset, we tested on a subset with 2000 randomly selected samples due to computation costs. 
\begin{figure}[ht!] 
  \centering 
  \setlength{\belowcaptionskip}{-10pt}
  \subfigure[\ourmethod for LSTM on the SST.]{ 
    \begin{minipage}{8 cm} 
      \centering 
      \includegraphics[height = 5 cm, width = 7.5 cm]{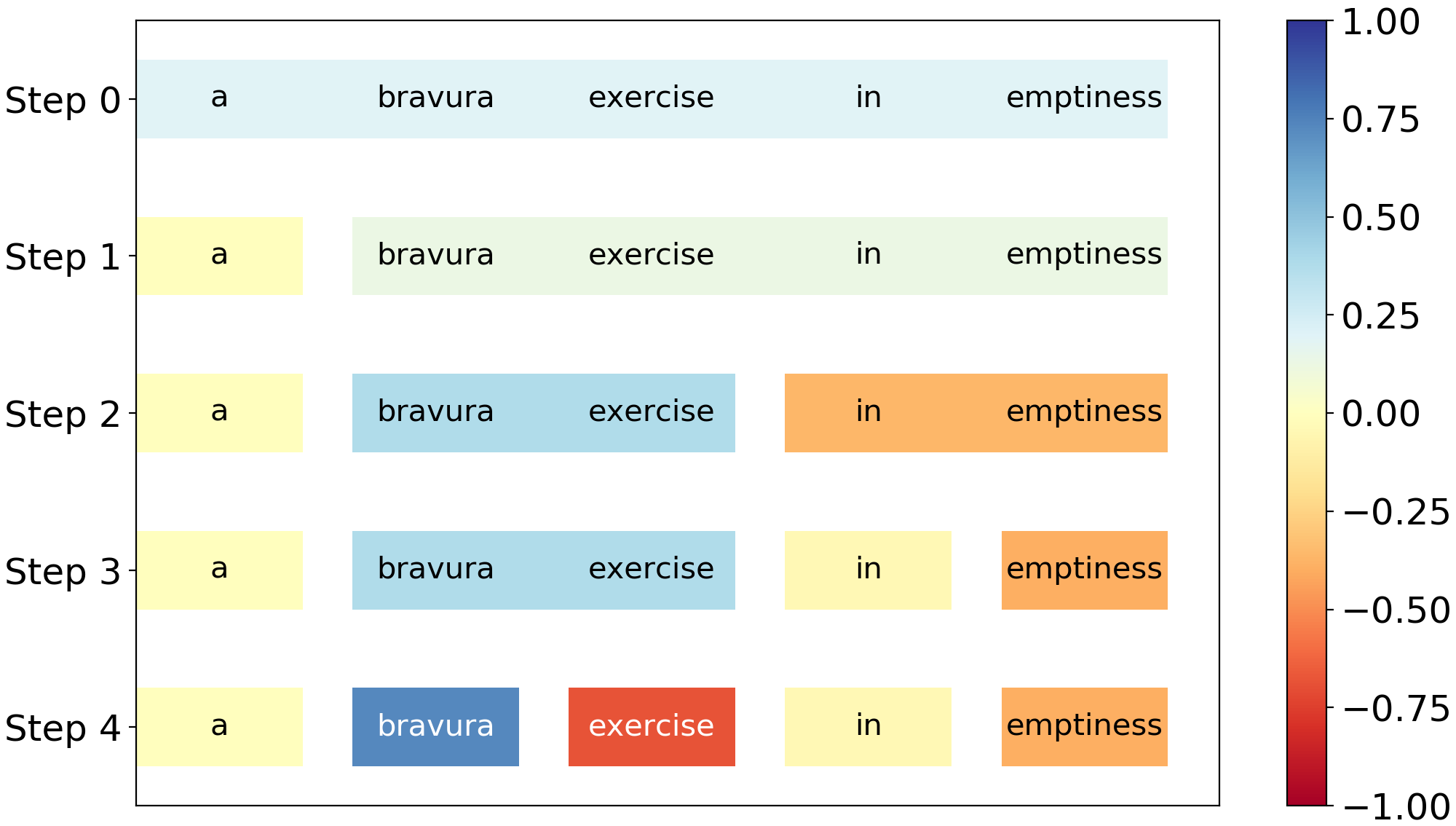} 
      \label{fig:lstm-intershap}
    \end{minipage} 
  } 
  \subfigure[ACD for LSTM on the SST.]{ 
    \begin{minipage}{8 cm} 
      \centering 
      \includegraphics[height = 5 cm, width = 7.5 cm]{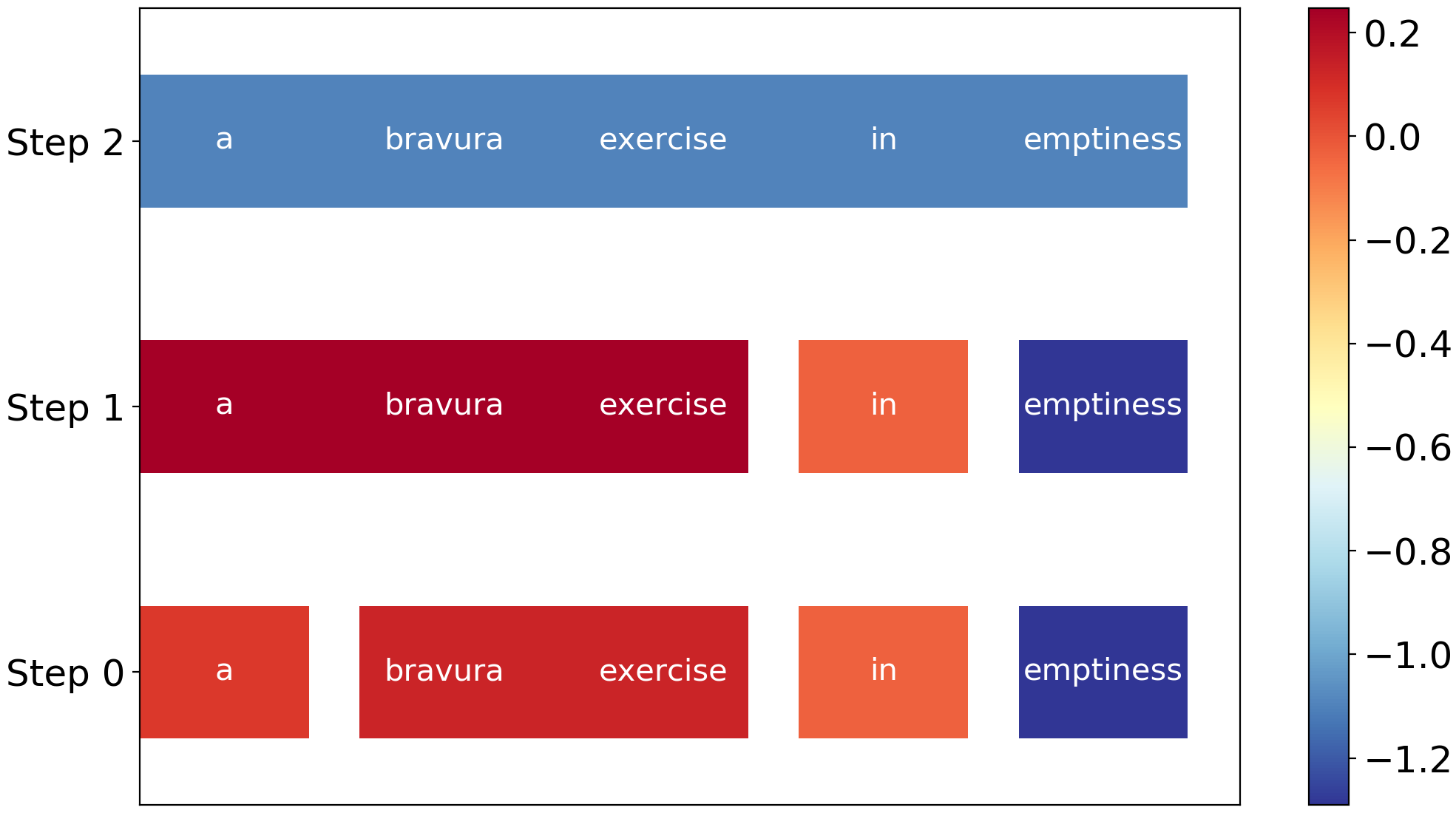}
      \label{fig:lstm-acd}
    \end{minipage} 
  } 
  \caption{Compare \ourmethod with ACD in interpreting the LSTM model on a negative movie review from the SST dataset, where LSTM makes a wrong prediction (\textsc{positive}). The importance scores of \ourmethod and CD scores are normalized for comparison.} 
  \label{fig:lstm-intershap-acd} 
\end{figure}
\ourmethod achieves the best performance on both evaluation metrics. 
SampleShapley also achieves a good performance with the number of samples set as 100, but the computational complexity is 200 times than \ourmethod.
Other variants, L/C-Shapley and KernelSHAP, applying approximations to Shapley values perform worse than SampleShapley and \ourmethod.
LIME performs comparatively to SampleShapley on the LSTM and CNN models, but is not fully capable of interpreting the deep neural network BERT. 
The limitation of context decomposition mentioned by \citet{jin2019hierarchical} is validated by the worst performance of CD in identifying important words. 
We also observed an interesting phenomenon that the simplest baseline Leave-one-out can achieve relatively good performance, even better than \ourmethod when $k$ and $r$ are small.
And we suspect that is because the criteria of Leave-one-out for picking single keywords matches the evaluation metrics. Overall, experimental results demonstrate the effectiveness of \autoref{eq:local-importance-score} in measuring feature importance. 
And the computational complexity is only $\mathcal{O}(n)$, which is much smaller than other baselines (e.g. SampleShapley, and L/C-Shapley with polynomial complexity).

\autoref{tab:phrase-level-evaluation} shows the cohesion-scores of \ourmethod and ACD with different models on the SST and IMDB datasets. \ourmethod outperforms ACD with LSTM, achieving higher cohesion-scores on both datasets, which indicates that \ourmethod is good at capturing important phrases. 
Comparing the results of \ourmethod on different models, the cohesion-scores of BERT are significantly higher than LSTM and CNN. 
It indicates that BERT is more sensitive to perturbations on important phrases and tends to utilize context information for predictions.

\subsection{Qualitative Analysis}
\label{subsec:qualiexp}

For qualitative analysis, we present two typical examples.
In the first example, we compare \ourmethod with ACD in interpreting the LSTM model.
\autoref{fig:lstm-intershap-acd} visualizes two hierarchical explanations, generated by \ourmethod and ACD respectively, on a negative movie review from the SST dataset. In this case, LSTM makes a wrong prediction (\textsc{positive}).
Figure \autoref{fig:lstm-intershap} shows \ourmethod correctly captures the sentiment polarities of \texttt{bravura} and \texttt{emptiness}, and the interaction between them as \texttt{bravura exercise} flips the polarity of \texttt{in emptiness} to positive. It explains why the model makes the wrong prediction. 
On the other hand, ACD incorrectly marks the two words with opposite polarities, and misses the feature interaction, as Figure \autoref{fig:lstm-acd} shows. 

\begin{figure}[ht!] 
  \centering 
  \setlength{\belowcaptionskip}{-10pt}
  \subfigure[\ourmethod for LSTM on SST.]{ 
    \begin{minipage}{8 cm} 
      \centering 
      \includegraphics[height = 5 cm, width = 7.5 cm]{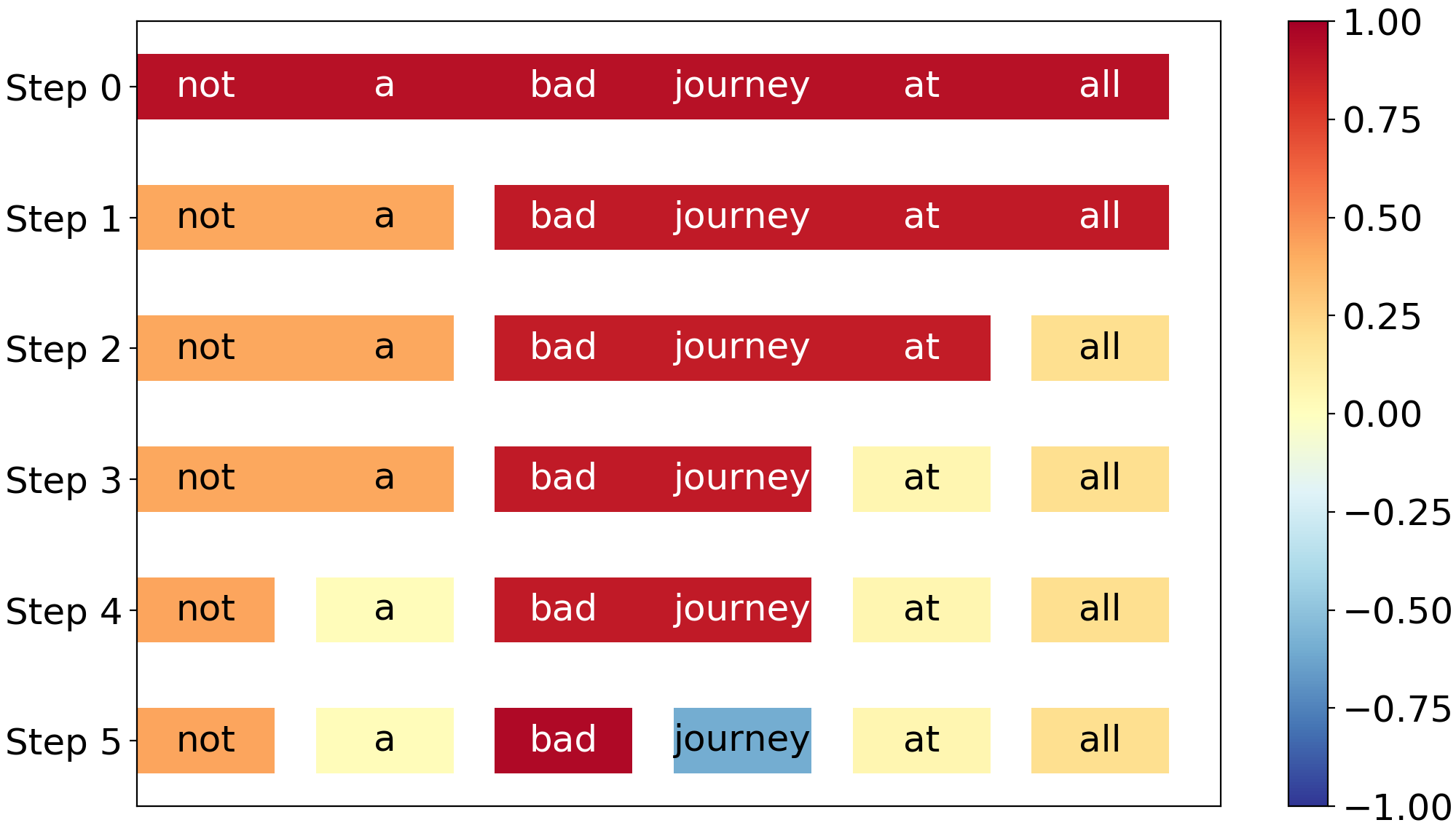} 
      \label{fig:lstm-not-a-bad}
   \end{minipage} 
  } 
  \subfigure[\ourmethod for BERT on SST.]{ 
   \begin{minipage}{8 cm} 
      \centering 
      \includegraphics[height = 5 cm, width = 7.5 cm]{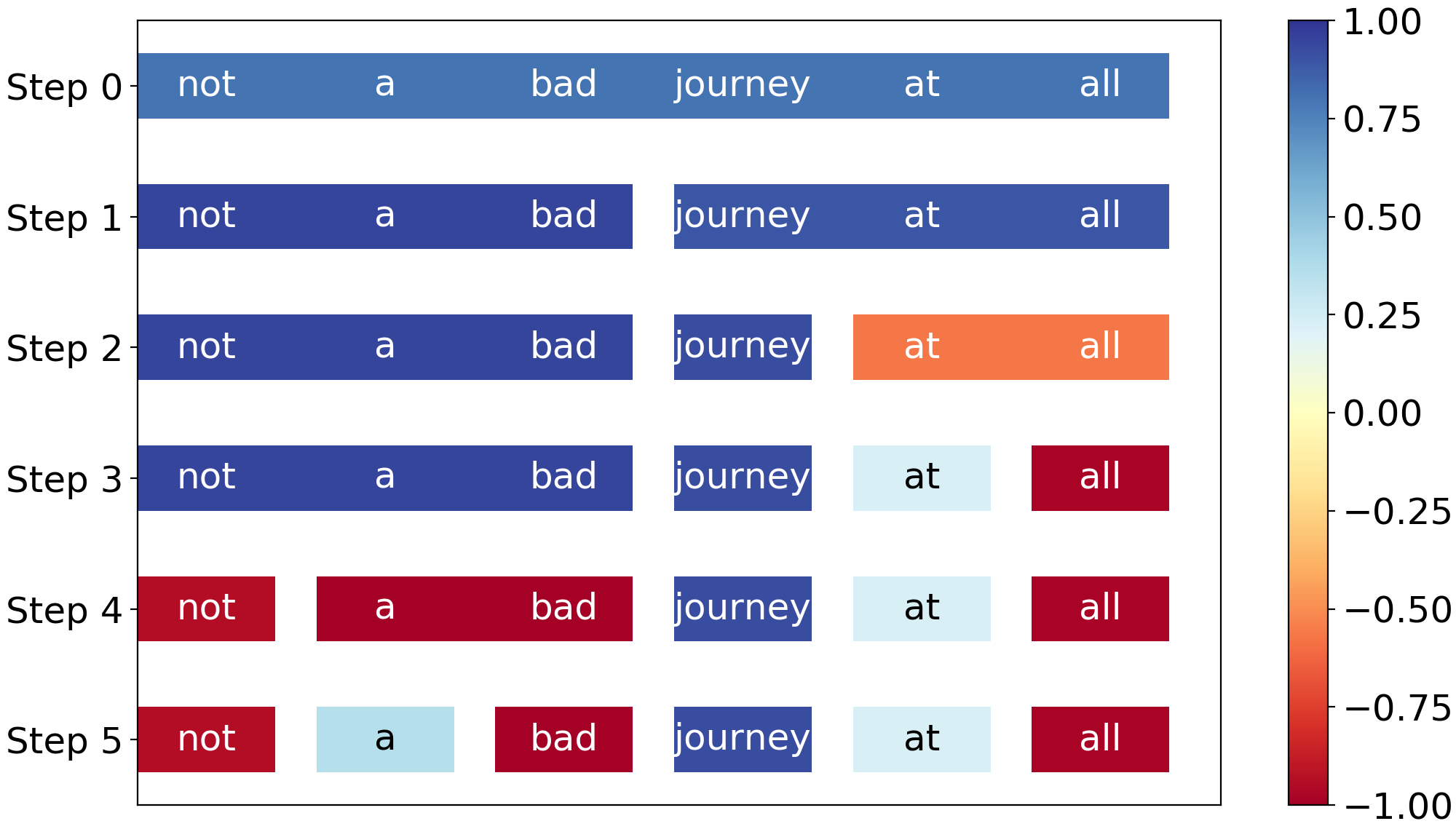}
      \label{fig:bert-not-a-bad}
   \end{minipage} 
  } 
  \caption{Compare \ourmethod in interpreting different models (LSTM and BERT) on a positive movie review from the SST dataset, where BERT makes the correct prediction (\textsc{positive}), while LSTM makes a wrong prediction (\textsc{negative}). \ourmethod explains that BERT captures the important phrase \texttt{not a bad} for making the correct prediction, while LSTM ignores it and is misled by the negative word \texttt{bad}.}
  \label{fig:bert-lstm-not-a-bad} 
\end{figure}

In the second example, we compare \ourmethod in interpreting two different models (LSTM and BERT).
\autoref{fig:bert-lstm-not-a-bad} visualizes the explanations on a positive movie review. 
In this case, BERT gives the correct prediction (\textsc{positive}), while LSTM makes a wrong prediction (\textsc{negative}). 
The comparison between Figure \autoref{fig:lstm-not-a-bad} and \autoref{fig:bert-not-a-bad} shows the difference of feature interactions within the two models and explains how a correct/wrong prediction was made.
Specifically, Figure~\ref{fig:bert-not-a-bad} illustrates that BERT captures the key phrase \texttt{not a bad} at step 1, and thus makes the positive prediction, while LSTM (as shown in Figure~\ref{fig:lstm-not-a-bad}) misses the interaction between \texttt{not} and \texttt{bad}, and the negative word \texttt{bad} pushes the model making the \textsc{negative} prediction. Both cases show that \ourmethod is capable of explaining model prediction behaviors, which helps humans understand the decision-making.
More examples are presented in \autoref{sec:visual_inter} due to the page limitation.

\subsection{Human Evaluation}
\label{subsec:humaneva}
We had 9 human annotators from the Amazon Mechanical Turk (AMT) for human evaluation. The features (e.g., words or phrases) with the highest importance score given by \ourmethod and other baselines are selected as the explanations. Note that \ourmethod and ACD can potentially give very long top features which are not user-friendly in human evaluation, so we additionally limit the maximum length of selected features to five. 
We provided the input text with different explanations in the user interface (as shown in Appendix~\ref{sec:interface}) and asked human annotators to guess the model's prediction~\citep{nguyen2018comparing} from \{``Negative", ``Positive", ``N/A"\} based on each explanation, where ``N/A" was selected when annotators cannot guess the model's prediction. 
We randomly picked 100 movie reviews from the IMDB dataset for human evaluation.

There are two dimensions of human evaluation.
We first compare \ourmethod with other baselines using the predictions made by the same LSTM model. 
Second, we compare the explanations generated by \ourmethod on three different models: LSTM, CNN, and BERT. 
We measure the number of human annotations that are \emph{coherent} with the actual model predictions, and define the \textit{coherence score} as the ratio between the coherent annotations and the total number of examples.

 

\subsubsection{Results}
\autoref{tab:humaneva} shows the coherence scores of eight different interpretation methods for LSTM on the IMDB dataset. \ourmethod outperforms other baselines with higher coherence score, which means that \ourmethod can capture important features which are highly consistent with human interpretations. LIME is still a strong baseline in providing interpretable explanations, while ACD and Shapley-based methods perform worse.
\autoref{tab:humaneva1} shows both the accuracy and coherence scores of different models. \ourmethod succeeds in interpreting black-box models with relatively high coherence scores. 
Moreover, although BERT can achieve higher prediction accuracy than the other two models, its coherence score is lower, manifesting a potential tradeoff between accuracy and interpretability of deep models.
\begin{table}[tbph]
  \centering
  \begin{tabular}{p{0.25\textwidth}c}
    	\toprule
         Methods & Coherence Score\\
         \midrule
         Leave-one-out &0.82 \\
          ACD & 0.68 \\
          LIME & 0.85\\
          L-Shapley &0.75 \\
          C-Shapley & 0.73\\
          KernelSHAP & 0.56\\
          SampleShapley &0.78 \\
         \ourmethod &\textbf{ 0.89} \\
         \bottomrule
  \end{tabular}
  \caption{\label{tab:humaneva}Human evaluation of different interpretation methods with LSTM model on the IMDB dataset.}
\end{table}%
\begin{table}[tbph]
  \centering
  \begin{tabular}{p{0.15\textwidth}cc}
    	\toprule
         Models & Accuracy & Coherence scores \\
         \midrule
         LSTM & 0.87 & 0.89 \\
         CNN & 0.90  & 0.84 \\
         BERT & 0.97 & 0.75 \\
         \bottomrule
  \end{tabular}
  \caption{\label{tab:humaneva1}Human evaluation of \ourmethod with different models on the IMDB dataset.}
\end{table}%

\section{Conclusion}
\label{sec:conclusion}
In this paper, we proposed an effective method, \ourmethod, building model-agnostic hierarchical interpretations via detecting feature interactions. In this work, we mainly focus on sentiment classification task. We test \ourmethod with three different neural network models on two benchmark datasets, and compare it with several competitive baseline methods. The superiority of \ourmethod is approved by both automatic and human evaluations.



\bibliography{ref}

\begin{thebibliography}{43}
\expandafter\ifx\csname natexlab\endcsname\relax\def\natexlab#1{#1}\fi

\bibitem[{Alvarez-Melis and Jaakkola(2018)}]{alvarez2018towards}
David Alvarez-Melis and Tommi~S Jaakkola. 2018.
\newblock Towards robust interpretability with self-explaining neural networks.
\newblock In \emph{NeurIPS}.

\bibitem[{Burns et~al.(2018)Burns, Nematzadeh, Grant, Gopnik, and
  Griffiths}]{burns2018exploiting}
Kaylee Burns, Aida Nematzadeh, Erin Grant, Alison Gopnik, and Tom Griffiths.
  2018.
\newblock Exploiting attention to reveal shortcomings in memory models.
\newblock In \emph{Proceedings of the 2018 EMNLP Workshop BlackboxNLP:
  Analyzing and Interpreting Neural Networks for NLP}, pages 378--380.

\bibitem[{Chen and Jordan(2019)}]{chen2019ls}
Jianbo Chen and Michael~I Jordan. 2019.
\newblock Ls-tree: Model interpretation when the data are linguistic.
\newblock \emph{arXiv preprint arXiv:1902.04187}.

\bibitem[{Chen et~al.(2018)Chen, Song, Wainwright, and
  Jordan}]{chen2018shapley}
Jianbo Chen, Le~Song, Martin~J Wainwright, and Michael~I Jordan. 2018.
\newblock L-shapley and c-shapley: Efficient model interpretation for
  structured data.
\newblock \emph{arXiv preprint arXiv:1808.02610}.

\bibitem[{Chen et~al.(2016)Chen, Zhu, Ling, Wei, Jiang, and
  Inkpen}]{chen2016enhanced}
Qian Chen, Xiaodan Zhu, Zhenhua Ling, Si~Wei, Hui Jiang, and Diana Inkpen.
  2016.
\newblock Enhanced lstm for natural language inference.
\newblock \emph{arXiv preprint arXiv:1609.06038}.

\bibitem[{Datta et~al.(2016)Datta, Sen, and Zick}]{datta2016algorithmic}
Anupam Datta, Shayak Sen, and Yair Zick. 2016.
\newblock Algorithmic transparency via quantitative input influence: Theory and
  experiments with learning systems.
\newblock In \emph{2016 IEEE symposium on security and privacy (SP)}, pages
  598--617. IEEE.

\bibitem[{Devlin et~al.(2018)Devlin, Chang, Lee, and
  Toutanova}]{devlin2018bert}
Jacob Devlin, Ming-Wei Chang, Kenton Lee, and Kristina Toutanova. 2018.
\newblock Bert: Pre-training of deep bidirectional transformers for language
  understanding.
\newblock \emph{arXiv preprint arXiv:1810.04805}.

\bibitem[{Fujimoto et~al.(2006)Fujimoto, Kojadinovic, and
  Marichal}]{fujimoto2006axiomatic}
Katsushige Fujimoto, Ivan Kojadinovic, and Jean-Luc Marichal. 2006.
\newblock Axiomatic characterizations of probabilistic and
  cardinal-probabilistic interaction indices.
\newblock \emph{Games and Economic Behavior}, 55(1):72--99.

\bibitem[{Ghaeini et~al.(2018)Ghaeini, Fern, and
  Tadepalli}]{ghaeini2018interpreting}
Reza Ghaeini, Xiaoli~Z Fern, and Prasad Tadepalli. 2018.
\newblock Interpreting recurrent and attention-based neural models: a case
  study on natural language inference.
\newblock \emph{arXiv preprint arXiv:1808.03894}.

\bibitem[{Godin et~al.(2018)Godin, Demuynck, Dambre, De~Neve, and
  Demeester}]{godin2018explaining}
Fr{\'e}deric Godin, Kris Demuynck, Joni Dambre, Wesley De~Neve, and Thomas
  Demeester. 2018.
\newblock Explaining character-aware neural networks for word-level prediction:
  Do they discover linguistic rules?
\newblock \emph{arXiv preprint arXiv:1808.09551}.

\bibitem[{Grabisch(1997)}]{grabisch1997k}
Michel Grabisch. 1997.
\newblock K-order additive discrete fuzzy measures and their representation.
\newblock \emph{Fuzzy sets and systems}, 92(2):167--189.

\bibitem[{Hechtlinger(2016)}]{hechtlinger2016interpretation}
Yotam Hechtlinger. 2016.
\newblock Interpretation of prediction models using the input gradient.
\newblock \emph{arXiv preprint arXiv:1611.07634}.

\bibitem[{Hochreiter and Schmidhuber(1997)}]{hochreiter1997long}
Sepp Hochreiter and J{\"u}rgen Schmidhuber. 1997.
\newblock Long short-term memory.
\newblock \emph{Neural computation}, 9(8):1735--1780.

\bibitem[{Howard and Ruder(2018)}]{howard2018universal}
Jeremy Howard and Sebastian Ruder. 2018.
\newblock Universal language model fine-tuning for text classification.
\newblock \emph{arXiv preprint arXiv:1801.06146}.

\bibitem[{Hu et~al.(2016)Hu, Xu, Rohrbach, Feng, Saenko, and
  Darrell}]{hu2016natural}
Ronghang Hu, Huazhe Xu, Marcus Rohrbach, Jiashi Feng, Kate Saenko, and Trevor
  Darrell. 2016.
\newblock Natural language object retrieval.
\newblock In \emph{Proceedings of the IEEE Conference on Computer Vision and
  Pattern Recognition}, pages 4555--4564.

\bibitem[{Jacovi et~al.(2018)Jacovi, Shalom, and
  Goldberg}]{jacovi2018understanding}
Alon Jacovi, Oren~Sar Shalom, and Yoav Goldberg. 2018.
\newblock Understanding convolutional neural networks for text classification.
\newblock \emph{arXiv preprint arXiv:1809.08037}.

\bibitem[{Jin et~al.(2019)Jin, Du, Wei, Xue, and Ren}]{jin2019hierarchical}
Xisen Jin, Junyi Du, Zhongyu Wei, Xiangyang Xue, and Xiang Ren. 2019.
\newblock \href {http://arxiv.org/abs/1911.06194} {Towards hierarchical
  importance attribution: Explaining compositional semantics for neural
  sequence models}.

\bibitem[{Jumelet and Hupkes(2018)}]{jumelet2018language}
Jaap Jumelet and Dieuwke Hupkes. 2018.
\newblock Do language models understand anything? on the ability of lstms to
  understand negative polarity items.
\newblock \emph{arXiv preprint arXiv:1808.10627}.

\bibitem[{Kim(2014)}]{kim2014convolutional}
Yoon Kim. 2014.
\newblock Convolutional neural networks for sentence classification.
\newblock \emph{arXiv preprint arXiv:1408.5882}.

\bibitem[{Kononenko et~al.(2010)}]{kononenko2010efficient}
Igor Kononenko et~al. 2010.
\newblock An efficient explanation of individual classifications using game
  theory.
\newblock \emph{Journal of Machine Learning Research}, 11(Jan):1--18.

\bibitem[{Lee et~al.(2017)Lee, Shin, and Kim}]{lee2017interactive}
Jaesong Lee, Joong-Hwi Shin, and Jun-Seok Kim. 2017.
\newblock Interactive visualization and manipulation of attention-based neural
  machine translation.
\newblock In \emph{Proceedings of the 2017 Conference on Empirical Methods in
  Natural Language Processing: System Demonstrations}, pages 121--126.

\bibitem[{Lei et~al.(2016)Lei, Barzilay, and Jaakkola}]{lei2016rationalizing}
Tao Lei, Regina Barzilay, and Tommi Jaakkola. 2016.
\newblock Rationalizing neural predictions.
\newblock In \emph{Proceedings of the 2016 Conference on Empirical Methods in
  Natural Language Processing}, pages 107--117.

\bibitem[{Li et~al.(2016)Li, Monroe, and Jurafsky}]{li2016understanding}
Jiwei Li, Will Monroe, and Dan Jurafsky. 2016.
\newblock Understanding neural networks through representation erasure.
\newblock \emph{arXiv preprint arXiv:1612.08220}.

\bibitem[{Lipton(2016)}]{lipton2016mythos}
Zachary~C Lipton. 2016.
\newblock The mythos of model interpretability.
\newblock \emph{arXiv preprint arXiv:1606.03490}.

\bibitem[{Lundberg et~al.(2018)Lundberg, Erion, and
  Lee}]{lundberg2018consistent}
Scott~M Lundberg, Gabriel~G Erion, and Su-In Lee. 2018.
\newblock Consistent individualized feature attribution for tree ensembles.
\newblock \emph{arXiv preprint arXiv:1802.03888}.

\bibitem[{Lundberg and Lee(2017)}]{lundberg2017unified}
Scott~M Lundberg and Su-In Lee. 2017.
\newblock A unified approach to interpreting model predictions.
\newblock In \emph{Advances in Neural Information Processing Systems}, pages
  4765--4774.

\bibitem[{Maas et~al.(2011)Maas, Daly, Pham, Huang, Ng, and
  Potts}]{maas2011learning}
Andrew~L Maas, Raymond~E Daly, Peter~T Pham, Dan Huang, Andrew~Y Ng, and
  Christopher Potts. 2011.
\newblock Learning word vectors for sentiment analysis.
\newblock In \emph{Proceedings of the 49th annual meeting of the association
  for computational linguistics: Human language technologies-volume 1}, pages
  142--150. Association for Computational Linguistics.

\bibitem[{Mikolov et~al.(2013)Mikolov, Sutskever, Chen, Corrado, and
  Dean}]{mikolov2013distributed}
Tomas Mikolov, Ilya Sutskever, Kai Chen, Greg~S Corrado, and Jeff Dean. 2013.
\newblock Distributed representations of words and phrases and their
  compositionality.
\newblock In \emph{Advances in neural information processing systems}, pages
  3111--3119.

\bibitem[{Murdoch et~al.(2018)Murdoch, Liu, and Yu}]{murdoch2018beyond}
W~James Murdoch, Peter~J Liu, and Bin Yu. 2018.
\newblock Beyond word importance: Contextual decomposition to extract
  interactions from lstms.
\newblock \emph{arXiv preprint arXiv:1801.05453}.

\bibitem[{Nguyen(2018)}]{nguyen2018comparing}
Dong Nguyen. 2018.
\newblock Comparing automatic and human evaluation of local explanations for
  text classification.
\newblock In \emph{Proceedings of the 2018 Conference of the North American
  Chapter of the Association for Computational Linguistics: Human Language
  Technologies, Volume 1 (Long Papers)}, pages 1069--1078.

\bibitem[{Owen(1972)}]{owen1972multilinear}
Guillermo Owen. 1972.
\newblock Multilinear extensions of games.
\newblock \emph{Management Science}, 18(5-part-2):64--79.

\bibitem[{Peters et~al.(2018)Peters, Neumann, Iyyer, Gardner, Clark, Lee, and
  Zettlemoyer}]{peters2018deep}
Matthew~E Peters, Mark Neumann, Mohit Iyyer, Matt Gardner, Christopher Clark,
  Kenton Lee, and Luke Zettlemoyer. 2018.
\newblock Deep contextualized word representations.
\newblock \emph{arXiv preprint arXiv:1802.05365}.

\bibitem[{Plumb et~al.(2018)Plumb, Molitor, and Talwalkar}]{plumb2018model}
Gregory Plumb, Denali Molitor, and Ameet~S Talwalkar. 2018.
\newblock Model agnostic supervised local explanations.
\newblock In \emph{Advances in Neural Information Processing Systems}, pages
  2515--2524.

\bibitem[{Ribeiro et~al.(2016)Ribeiro, Singh, and Guestrin}]{ribeiro2016should}
Marco~Tulio Ribeiro, Sameer Singh, and Carlos Guestrin. 2016.
\newblock Why should i trust you?: Explaining the predictions of any
  classifier.
\newblock In \emph{Proceedings of the 22nd ACM SIGKDD international conference
  on knowledge discovery and data mining}, pages 1135--1144. ACM.

\bibitem[{Samek et~al.(2016)Samek, Binder, Montavon, Lapuschkin, and
  M{\"u}ller}]{samek2016evaluating}
Wojciech Samek, Alexander Binder, Gr{\'e}goire Montavon, Sebastian Lapuschkin,
  and Klaus-Robert M{\"u}ller. 2016.
\newblock Evaluating the visualization of what a deep neural network has
  learned.
\newblock \emph{IEEE transactions on neural networks and learning systems},
  28(11):2660--2673.

\bibitem[{Serrano and Smith(2019)}]{serrano2019attention}
Sofia Serrano and Noah~A Smith. 2019.
\newblock Is attention interpretable?
\newblock \emph{arXiv preprint arXiv:1906.03731}.

\bibitem[{Shapley(1953)}]{shapley1953value}
Lloyd~S Shapley. 1953.
\newblock A value for n-person games.
\newblock \emph{Contributions to the Theory of Games}, 2(28).

\bibitem[{Shrikumar et~al.(2017)Shrikumar, Greenside, and
  Kundaje}]{shrikumar2017learning}
Avanti Shrikumar, Peyton Greenside, and Anshul Kundaje. 2017.
\newblock Learning important features through propagating activation
  differences.
\newblock In \emph{Proceedings of the 34th International Conference on Machine
  Learning-Volume 70}, pages 3145--3153. JMLR. org.

\bibitem[{Singh et~al.(2019)Singh, Murdoch, and Yu}]{singh2018hierarchical}
Chandan Singh, W.~James Murdoch, and Bin Yu. 2019.
\newblock \href {https://openreview.net/forum?id=SkEqro0ctQ} {Hierarchical
  interpretations for neural network predictions}.
\newblock In \emph{International Conference on Learning Representations}.

\bibitem[{Socher et~al.(2013)Socher, Perelygin, Wu, Chuang, Manning, Ng, and
  Potts}]{socher2013recursive}
Richard Socher, Alex Perelygin, Jean Wu, Jason Chuang, Christopher~D Manning,
  Andrew Ng, and Christopher Potts. 2013.
\newblock Recursive deep models for semantic compositionality over a sentiment
  treebank.
\newblock In \emph{Proceedings of the 2013 conference on empirical methods in
  natural language processing}, pages 1631--1642.

\bibitem[{{\v{S}}trumbelj and Kononenko(2014)}]{vstrumbelj2014explaining}
Erik {\v{S}}trumbelj and Igor Kononenko. 2014.
\newblock Explaining prediction models and individual predictions with feature
  contributions.
\newblock \emph{Knowledge and information systems}, 41(3):647--665.

\bibitem[{Sundararajan et~al.(2017)Sundararajan, Taly, and
  Yan}]{sundararajan2017axiomatic}
Mukund Sundararajan, Ankur Taly, and Qiqi Yan. 2017.
\newblock Axiomatic attribution for deep networks.
\newblock In \emph{Proceedings of the 34th International Conference on Machine
  Learning-Volume 70}, pages 3319--3328. JMLR. org.

\bibitem[{Tsang et~al.(2018)Tsang, Sun, Ren, and Liu}]{tsang2018can}
Michael Tsang, Youbang Sun, Dongxu Ren, and Yan Liu. 2018.
\newblock Can i trust you more? model-agnostic hierarchical explanations.
\newblock \emph{arXiv preprint arXiv:1812.04801}.

\end{thebibliography}
\bibliographystyle{acl_natbib}

\newpage
\appendix

\section{Comparison between Top-down and Bottom-up Approaches}
\label{sec:compare-top-bottom}
\begin{figure}[ht!] 
	\centering 
	\subfigure[Bottom-up clustering.]{ 
		\begin{minipage}{8 cm} 
			\centering 
			\includegraphics[height = 5 cm, width = 8 cm]{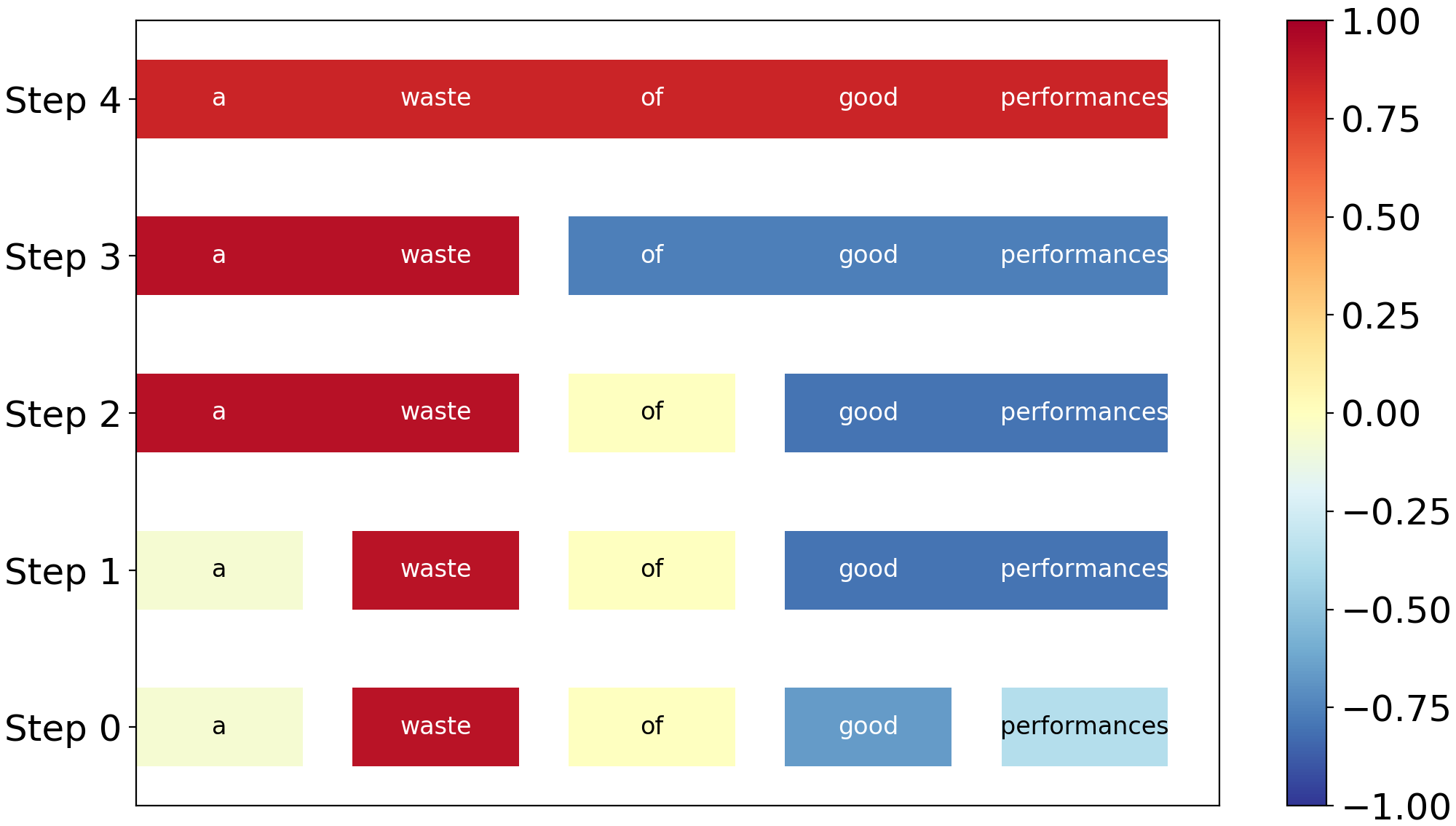} 
			\label{bottom-up}
		\end{minipage} 
	} 
	\subfigure[Top-down segmentation.]{ 
		\begin{minipage}{8 cm} 
			\centering 
			\includegraphics[height = 5 cm, width = 8 cm]{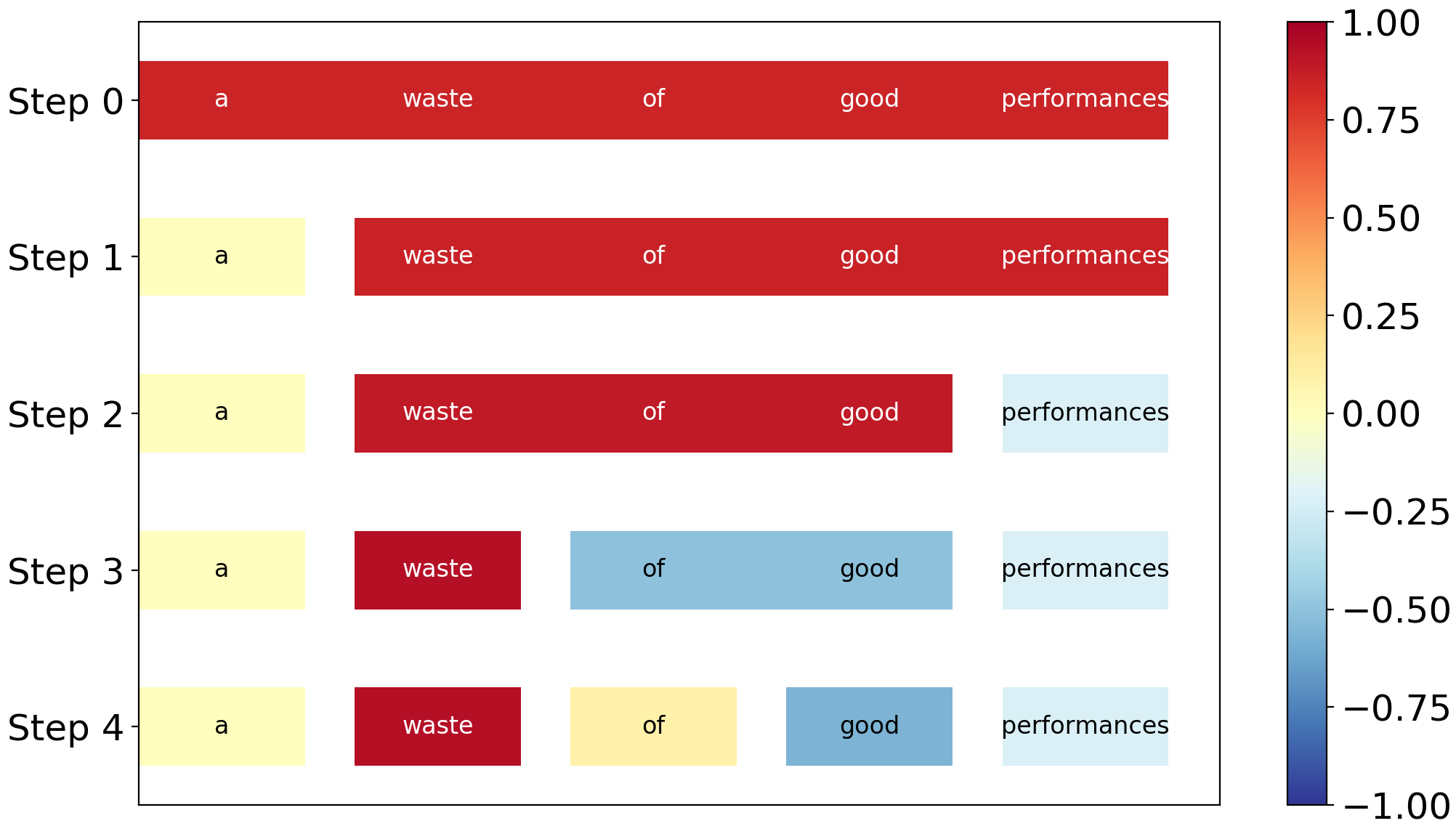}
			\label{top-down}
		\end{minipage} 
	} 
	\caption{Hierarchical interpretations for the LSTM model using the bottom-up and top-down approaches respectively. Red and green colors represent the negative and positive sentiments respectively.} 
		\label{fig:bottom-up-top-down-cmp} 
	\end{figure}

Given the sentence \texttt{a waste of good performance} for example, Figure~\ref{fig:bottom-up-top-down-cmp} shows the hierarchical interpretations for the LSTM model using the bottom-up and top-down approaches respectively. Figure~\ref{bottom-up} shows that the interaction between \texttt{waste} and \texttt{good} can not be captured until the last (top) layer, while the important phrase \texttt{waste of good} can be extracted in the intermediate layer by top-down algorithm. We can see that \texttt{waste} flips the polarity of \texttt{of good} to negative, causing the model predicting negative as well. Top-down segmentation performs better than bottom-up in capturing feature interactions. The reason is that the bottom layer contains more features than the top layer, which incurs larger errors in calculating interaction scores. Even worse, the calculation error will propagate and accumulate during clustering.

\section{Results of AOPCs and log-odds changing with different $k$ and $r$}
\label{sec:other_aopc_log}

\begin{figure}[H] 
	\centering 
	\subfigure[AOPCs of LSTM on the SST dataset.]{
	\begin{minipage}{8 cm} 
			\centering 
			\includegraphics[height = 6 cm, width = 8 cm]{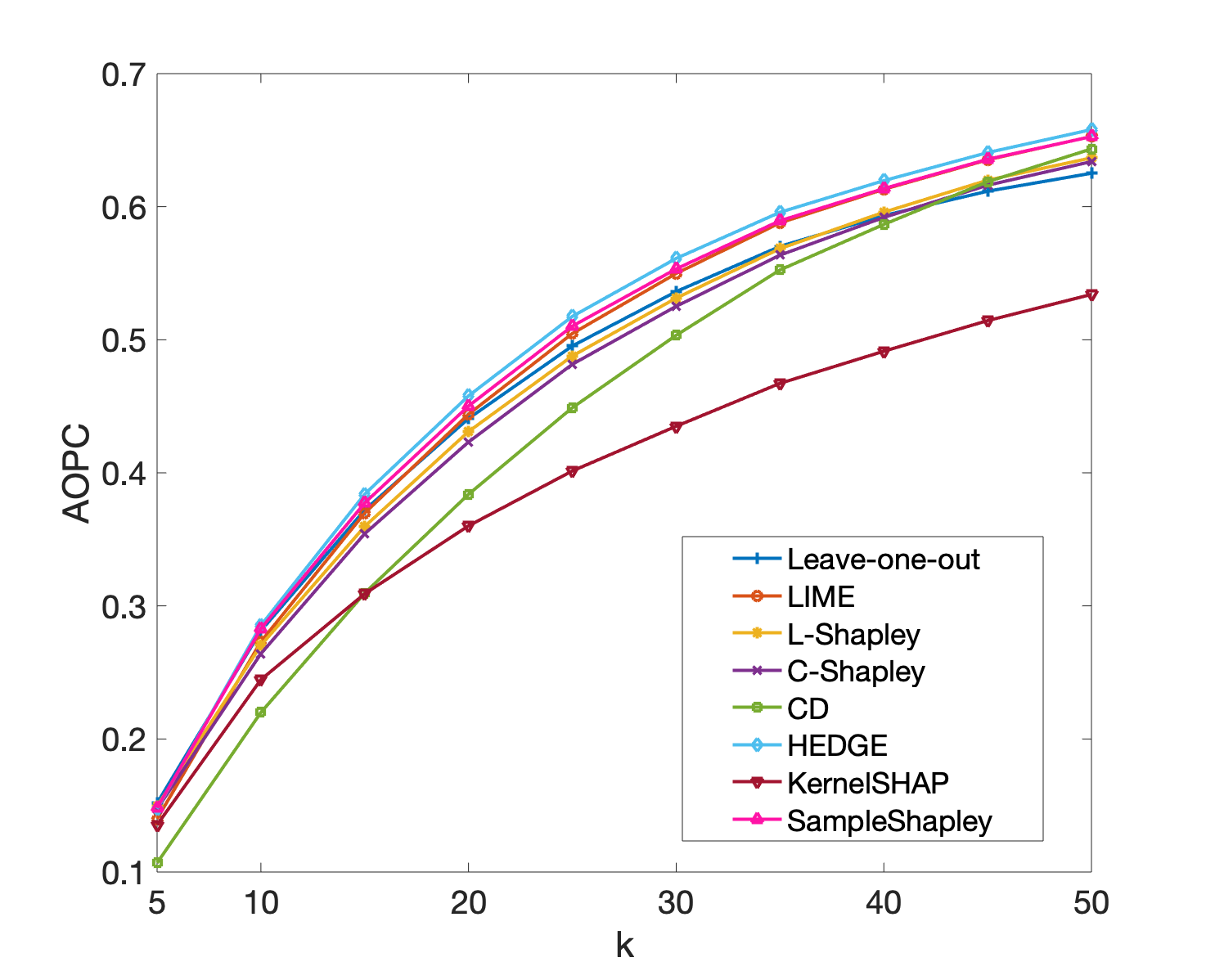} 
			\label{aopc_lstm_sst}
	\end{minipage} 
	}
	\subfigure[Log-odds of LSTM on the SST dataset.]{ 
		\begin{minipage}{8 cm} 
			\centering 
			\includegraphics[height = 6 cm, width = 8 cm]{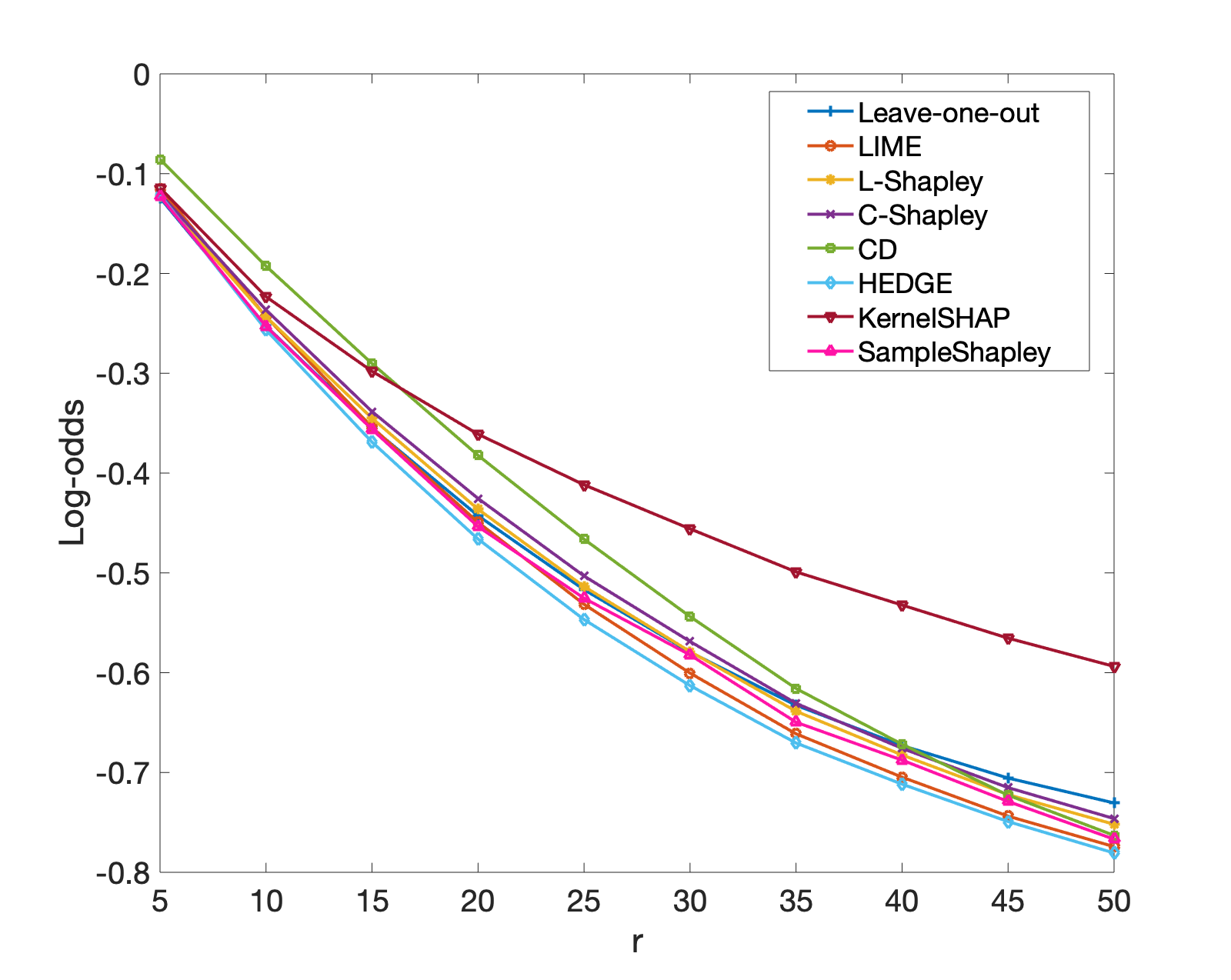}
			\label{log_lstm_sst}
		\end{minipage} 
	}
	\caption{The AOPC and log-odds for LSTM on the SST dataset.} 
	\label{fig:aopc_log_lstm_sst} 
\end{figure}

\begin{figure}[H] 
	\centering 
	\subfigure[AOPCs of LSTM on the IMDB dataset.]{
	\begin{minipage}{8 cm} 
			\centering 
			\includegraphics[height = 6 cm, width = 8 cm]{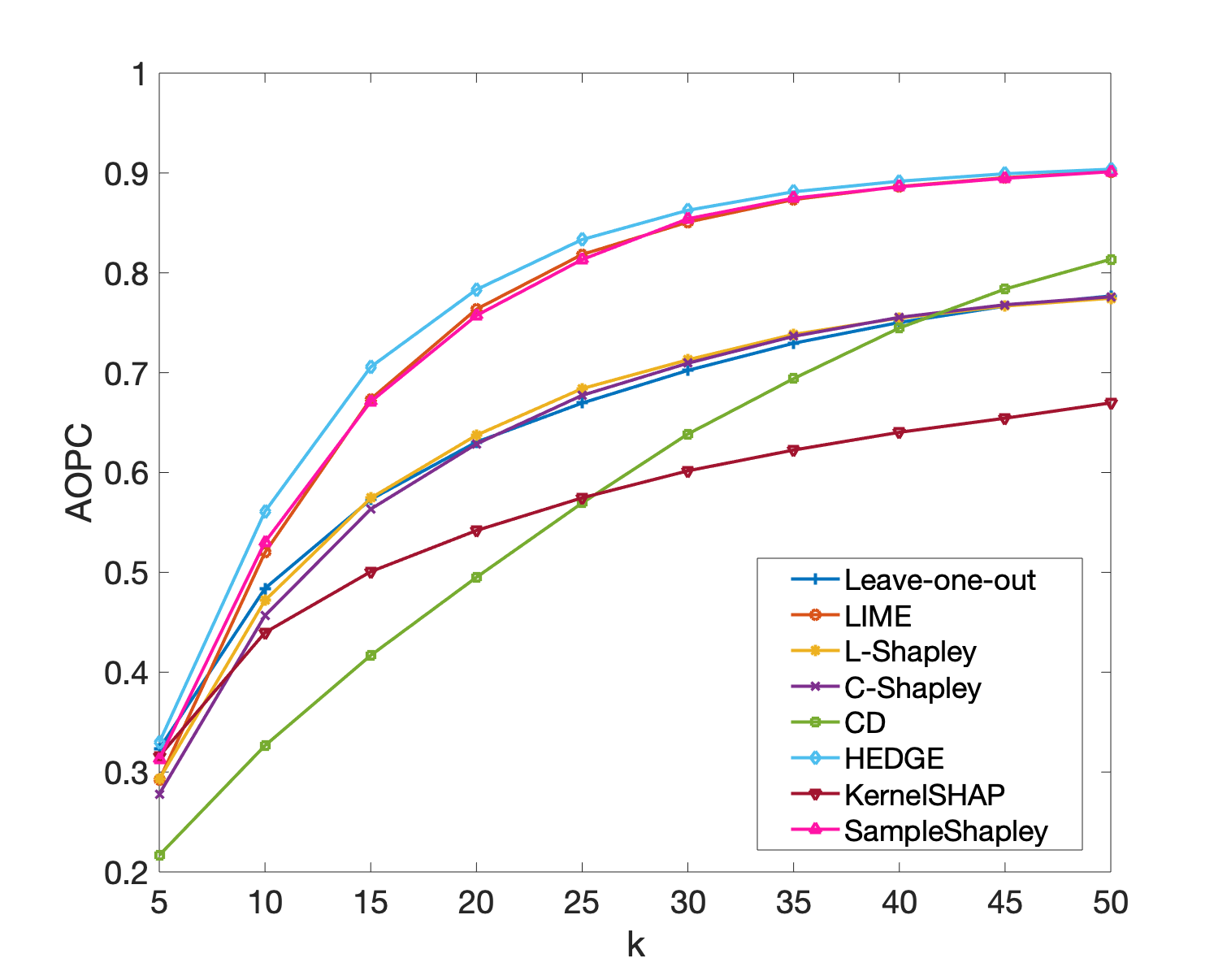} 
			\label{aopc_lstm_imdb}
	\end{minipage} 
	}
	\subfigure[Log-odds of LSTM on the IMDB dataset.]{ 
		\begin{minipage}{8 cm} 
			\centering 
			\includegraphics[height = 6 cm, width = 8 cm]{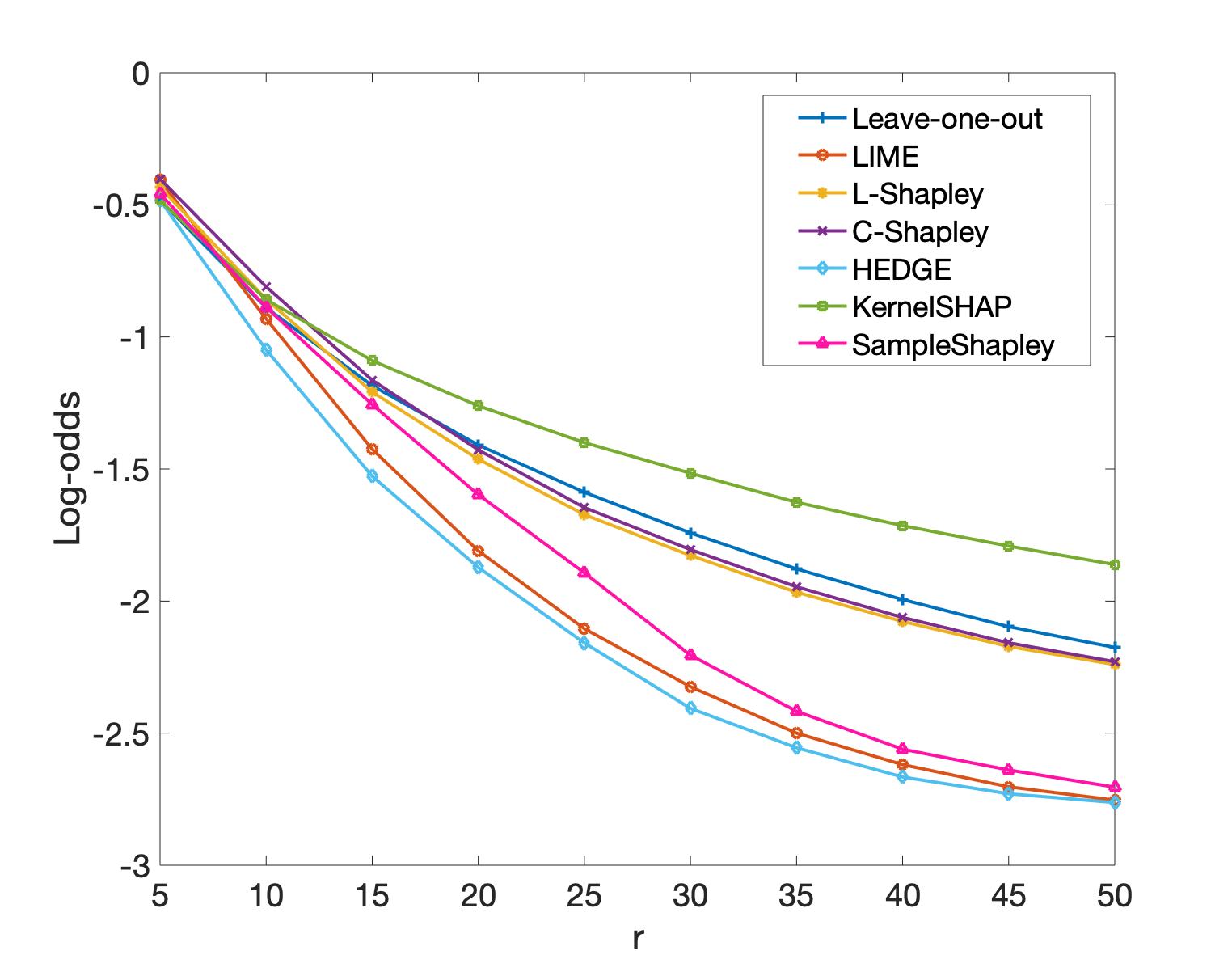}
			\label{log_lstm_imdb}
		\end{minipage} 
	}
	\caption{The AOPC and log-odds for LSTM on the IMDB dataset.} 
	\label{fig:aopc_log_lstm_imdb} 
\end{figure}

\begin{figure}[H] 
	\centering 
	\subfigure[AOPCs of CNN on the SST dataset.]{
	\begin{minipage}{8 cm} 
			\centering 
			\includegraphics[height = 6 cm, width = 8 cm]{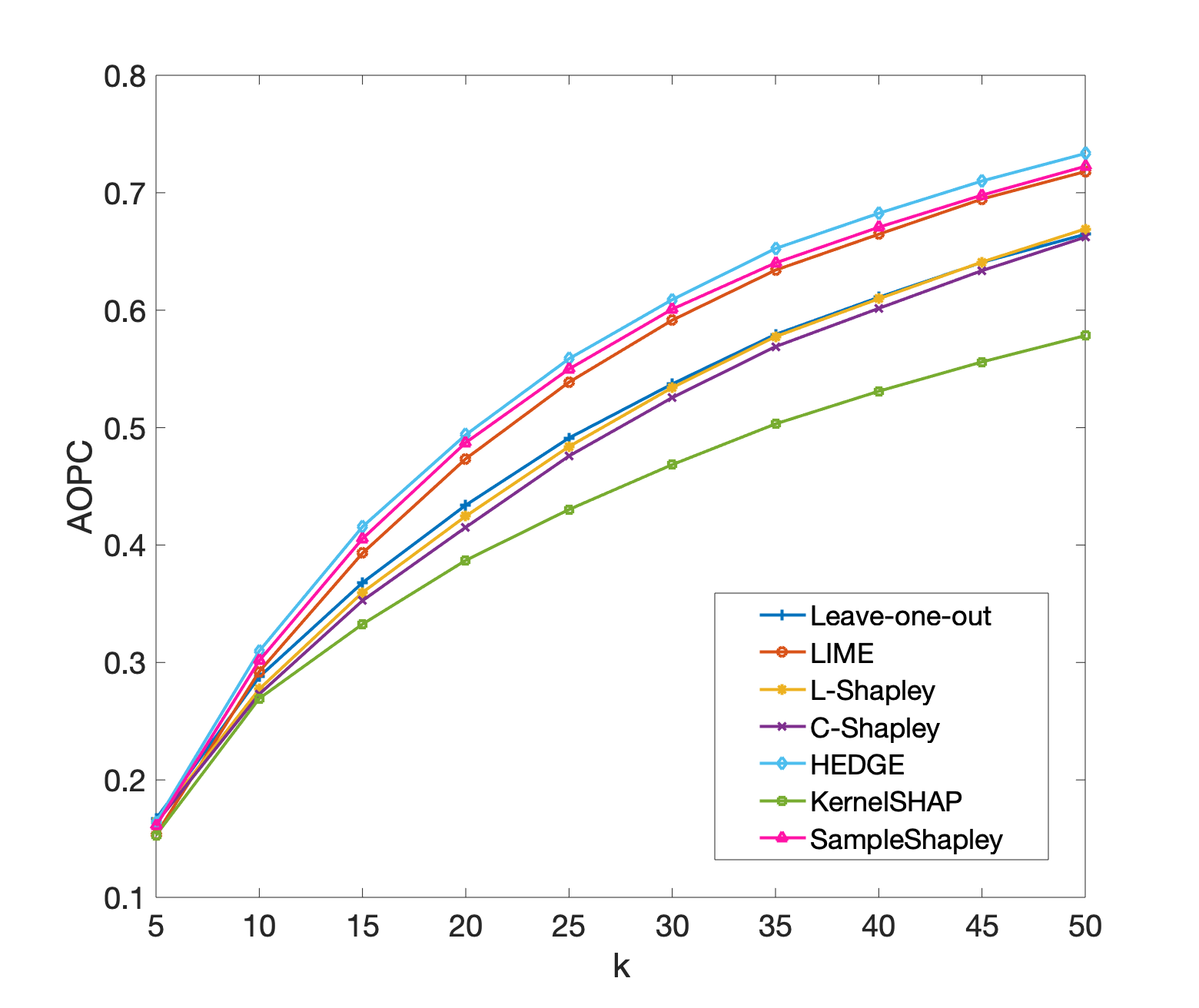} 
			\label{aopc_cnn_sst}
	\end{minipage} 
	}
	\subfigure[Log-odds of CNN on the SST dataset.]{ 
		\begin{minipage}{8 cm} 
			\centering 
			\includegraphics[height = 6 cm, width = 8 cm]{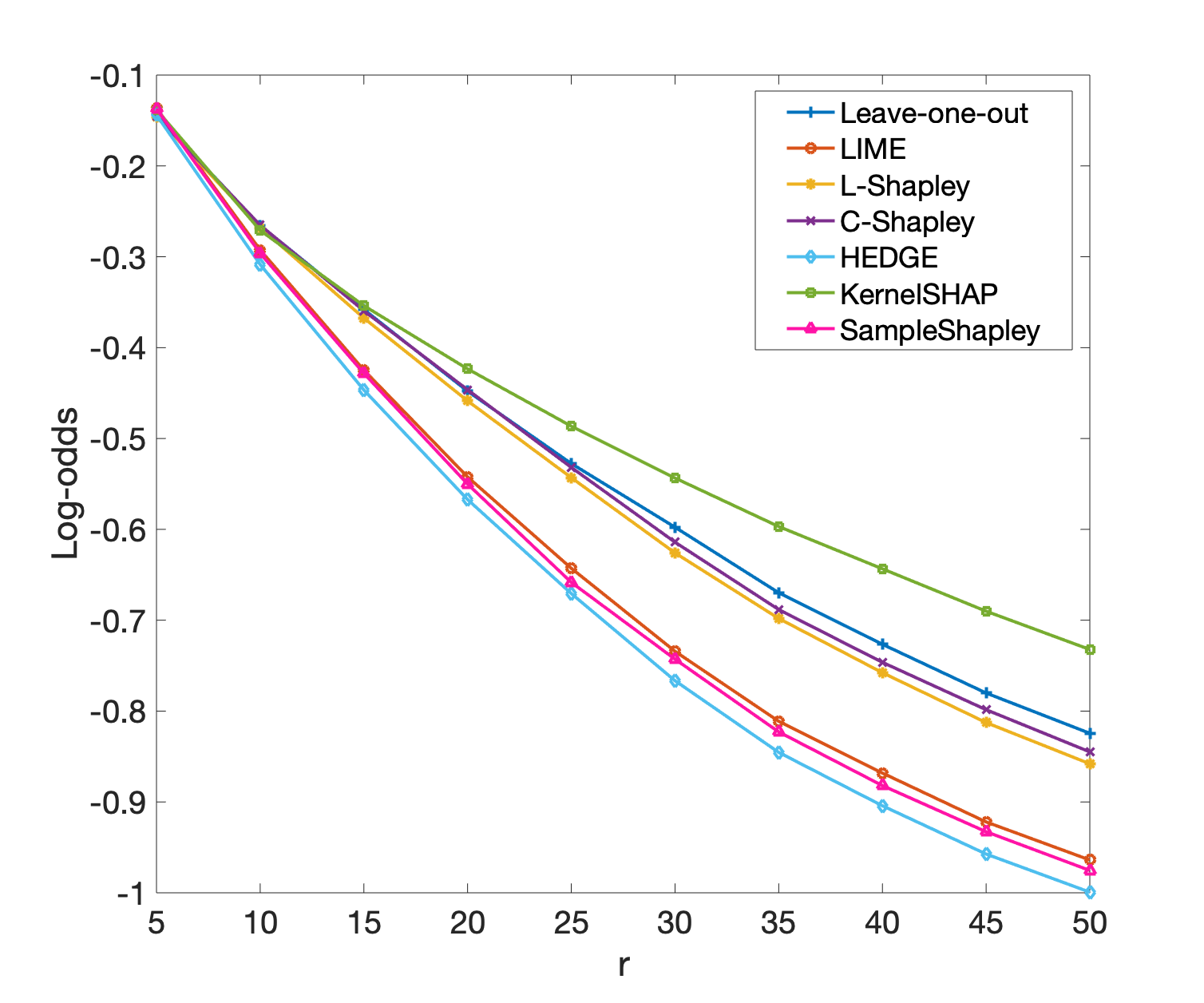}
			\label{log_cnn_sst}
		\end{minipage} 
	}
	\caption{The AOPC and log-odds for CNN on the SST dataset.} 
	\label{fig:aopc_log_cnn_sst} 
\end{figure}

\begin{figure}[H] 
	\centering 
	\subfigure[AOPCs of CNN on the IMDB dataset.]{
	\begin{minipage}{8 cm} 
			\centering 
			\includegraphics[height = 6 cm, width = 8 cm]{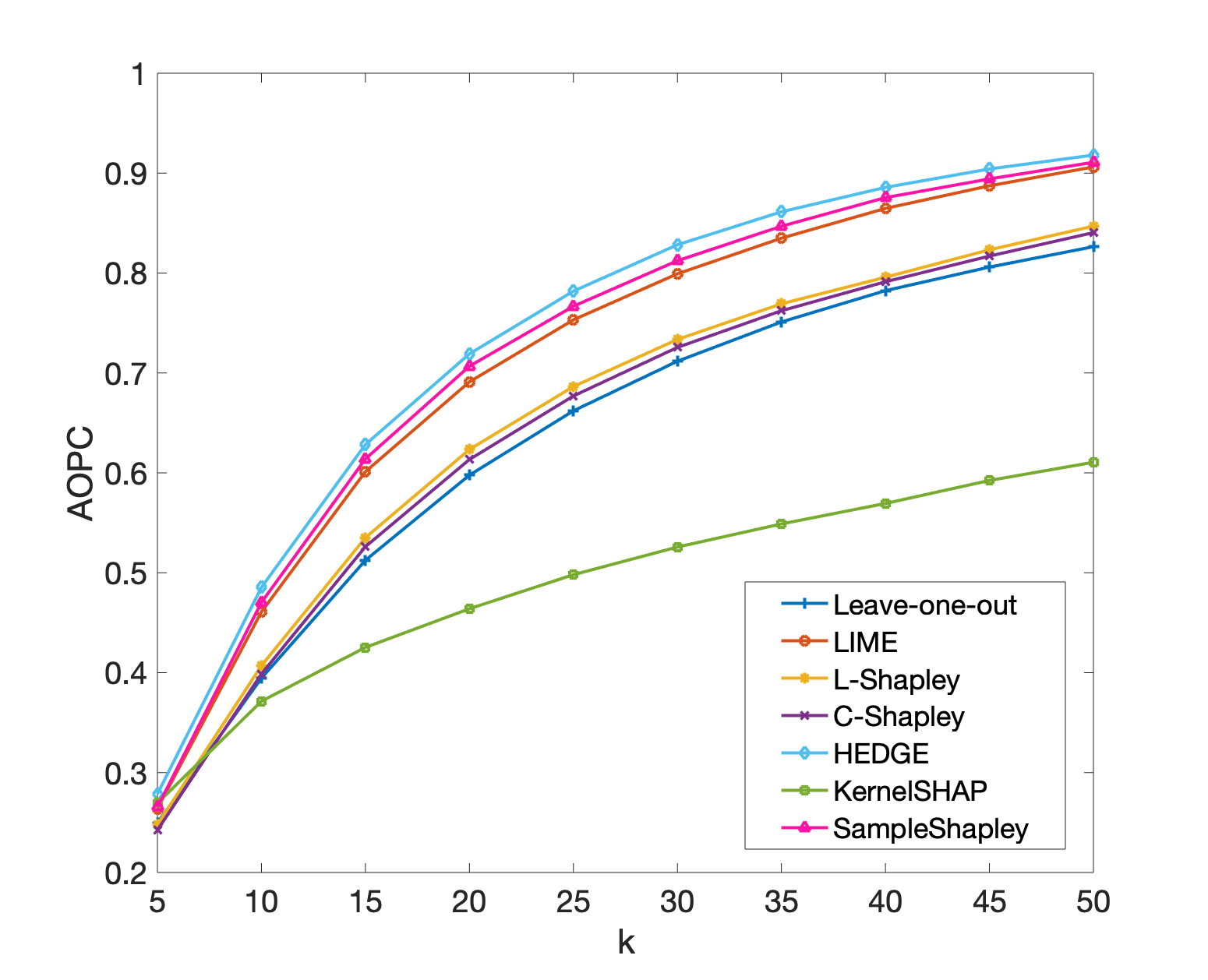} 
			\label{aopc_cnn_imdb}
	\end{minipage} 
	}
	\subfigure[Log-odds of CNN on the IMDB dataset.]{ 
		\begin{minipage}{8 cm} 
			\centering 
			\includegraphics[height = 6 cm, width = 8 cm]{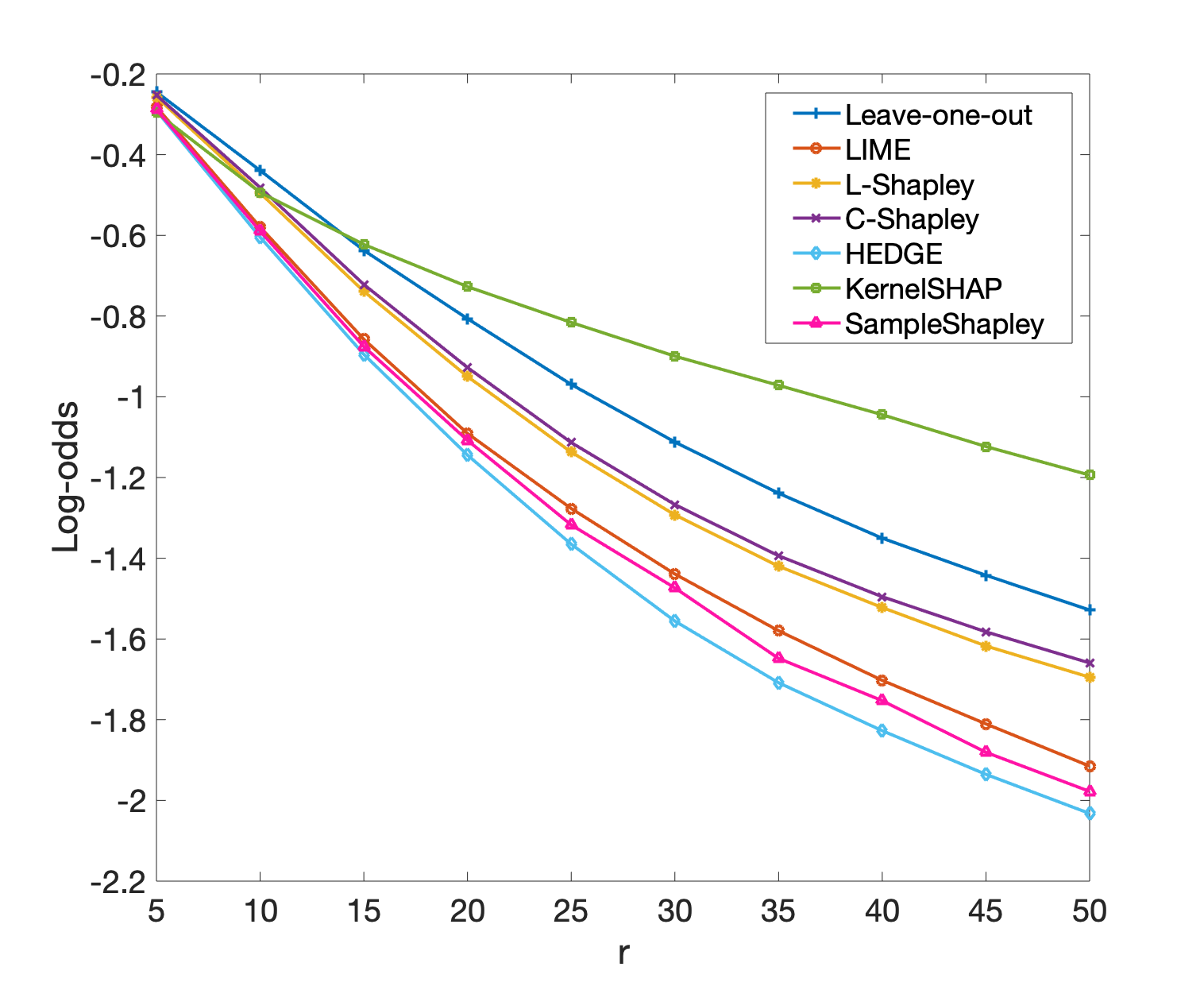}
			\label{log_cnn_imdb}
		\end{minipage} 
	}
	\caption{The AOPC and log-odds for CNN on the IMDB dataset.} 
	\label{fig:aopc_log_cnn_imdb} 
\end{figure}

\begin{figure}[H] 
	\centering 
	\subfigure[AOPCs of BERT on the SST dataset.]{
	\begin{minipage}{8 cm} 
			\centering 
			\includegraphics[height = 6 cm, width = 8 cm]{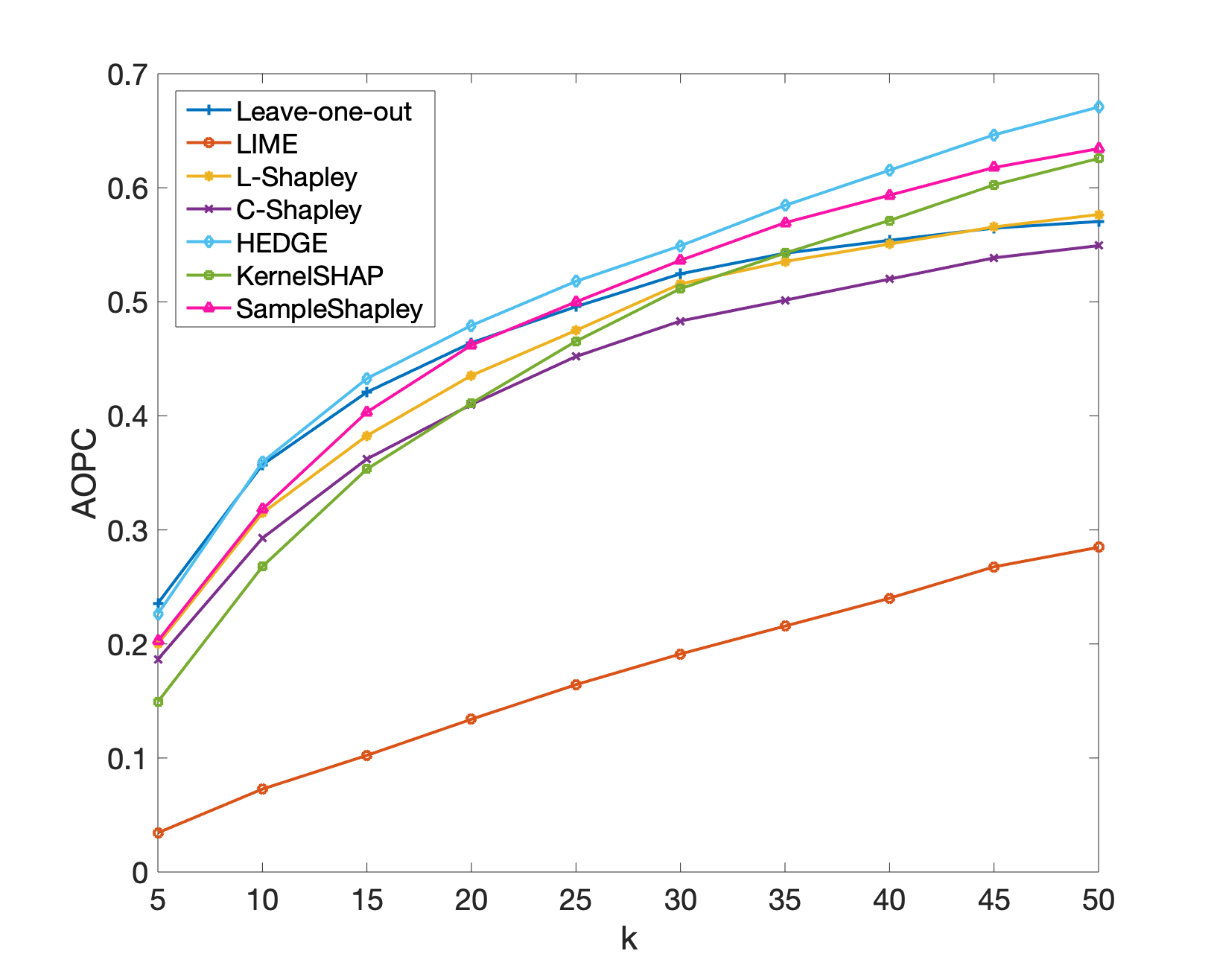} 
			\label{aopc_bert_sst}
	\end{minipage} 
	}
	\subfigure[Log-odds of BERT on the SST dataset.]{ 
		\begin{minipage}{8 cm} 
			\centering 
			\includegraphics[height = 6 cm, width = 8 cm]{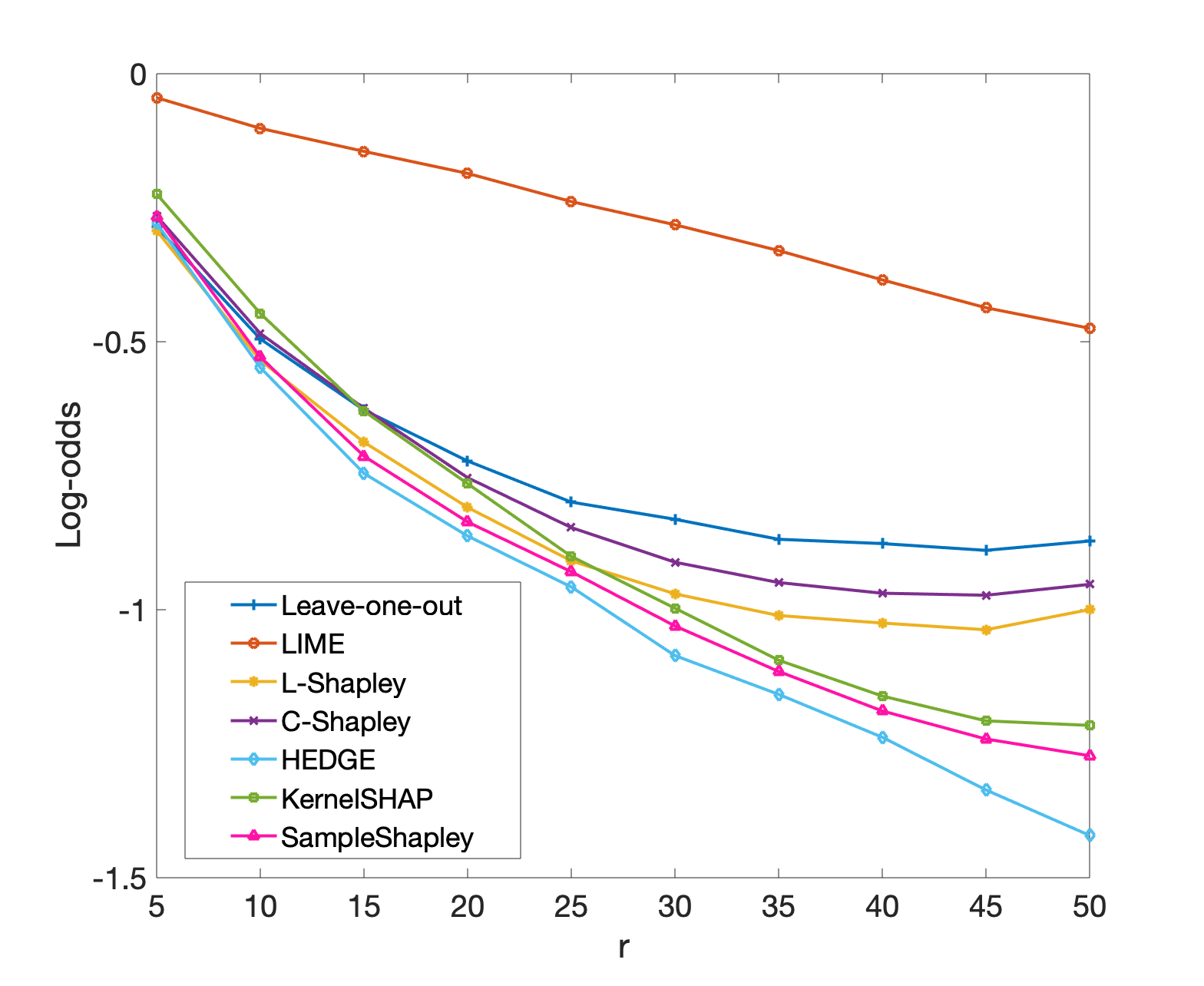}
			\label{log_bert_sst}
		\end{minipage} 
	}
	\caption{The AOPC and log-odds for BERT on the SST dataset.} 
	\label{fig:aopc_log_bert_sst} 
\end{figure}

\begin{figure}[H] 
	\centering 
	\subfigure[AOPCs of BERT on the IMDB dataset.]{
	\begin{minipage}{8 cm} 
			\centering 
			\includegraphics[height = 6 cm, width = 8 cm]{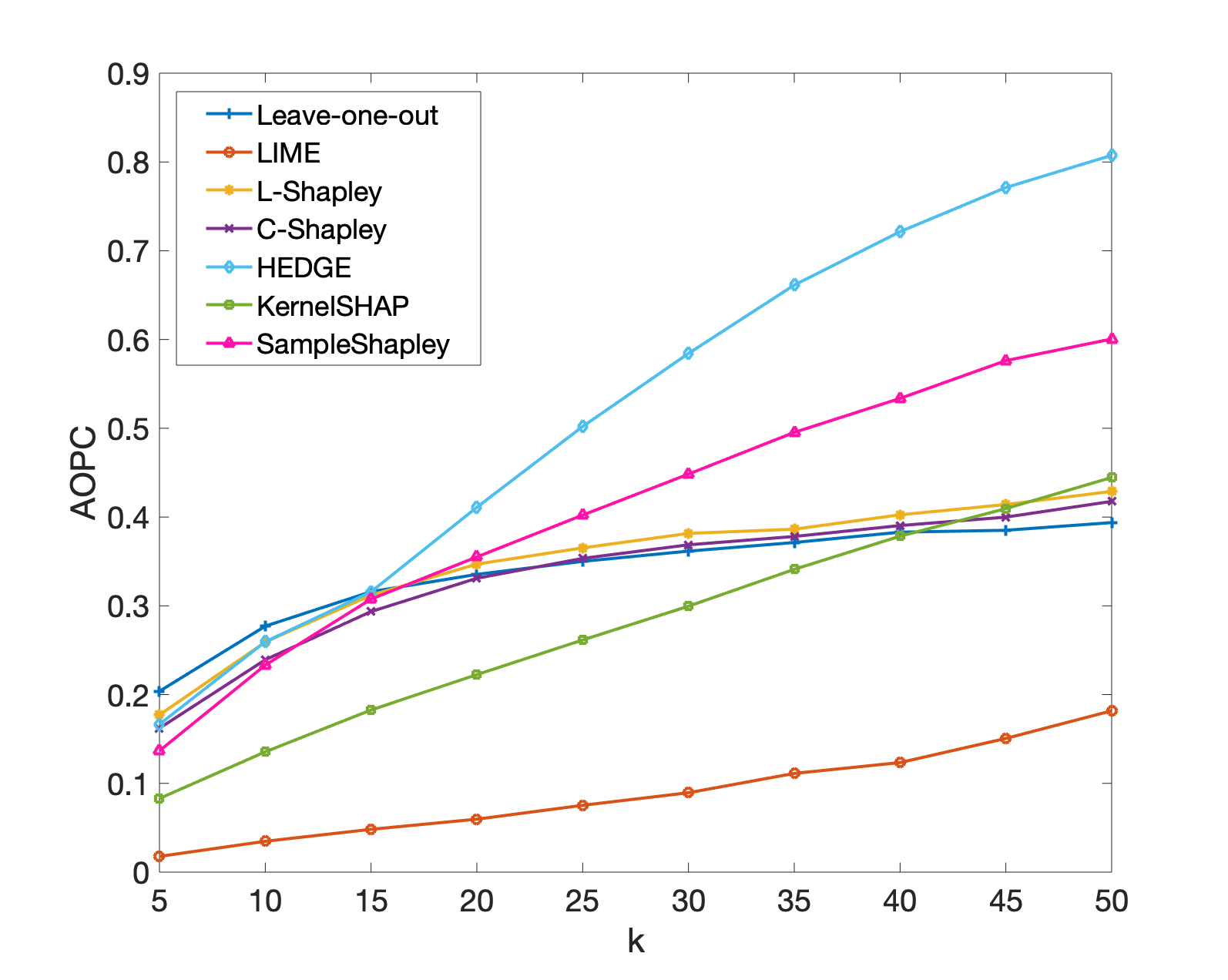} 
			\label{aopc_bert_imdb}
	\end{minipage} 
	}
	\subfigure[Log-odds of BERT on the IMDB dataset.]{ 
		\begin{minipage}{8 cm} 
			\centering 
			\includegraphics[height = 6 cm, width = 8 cm]{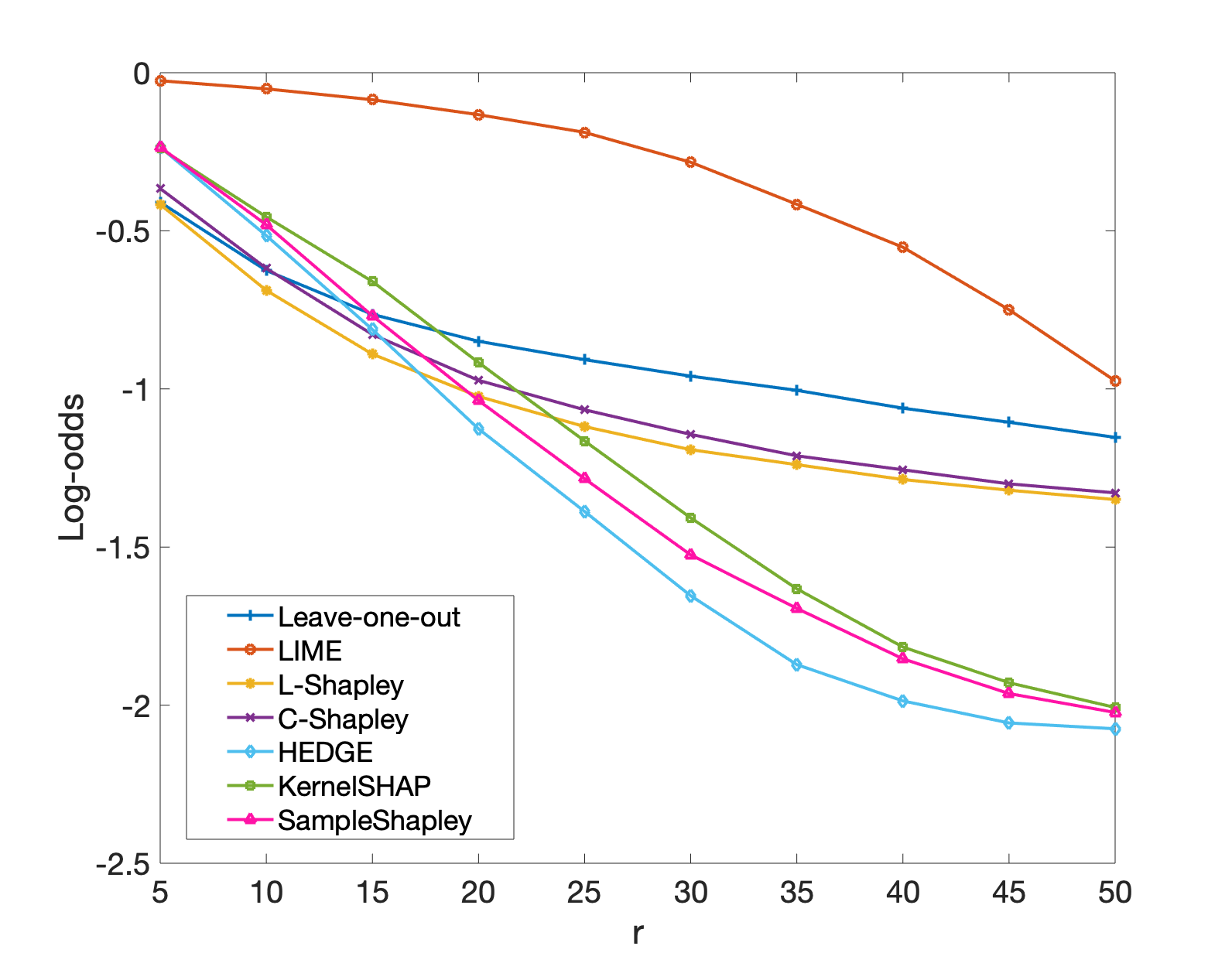}
			\label{log_bert_imdb}
		\end{minipage} 
	}
	\caption{The AOPC and log-odds for BERT on the IMDB dataset.} 
	\label{fig:aopc_log_bert_imdb} 
\end{figure}

\section{Visualization of Hierarchical Interpretations}
\label{sec:visual_inter}
\begin{figure}[H]        
  \center{\includegraphics[width=7.5cm]  {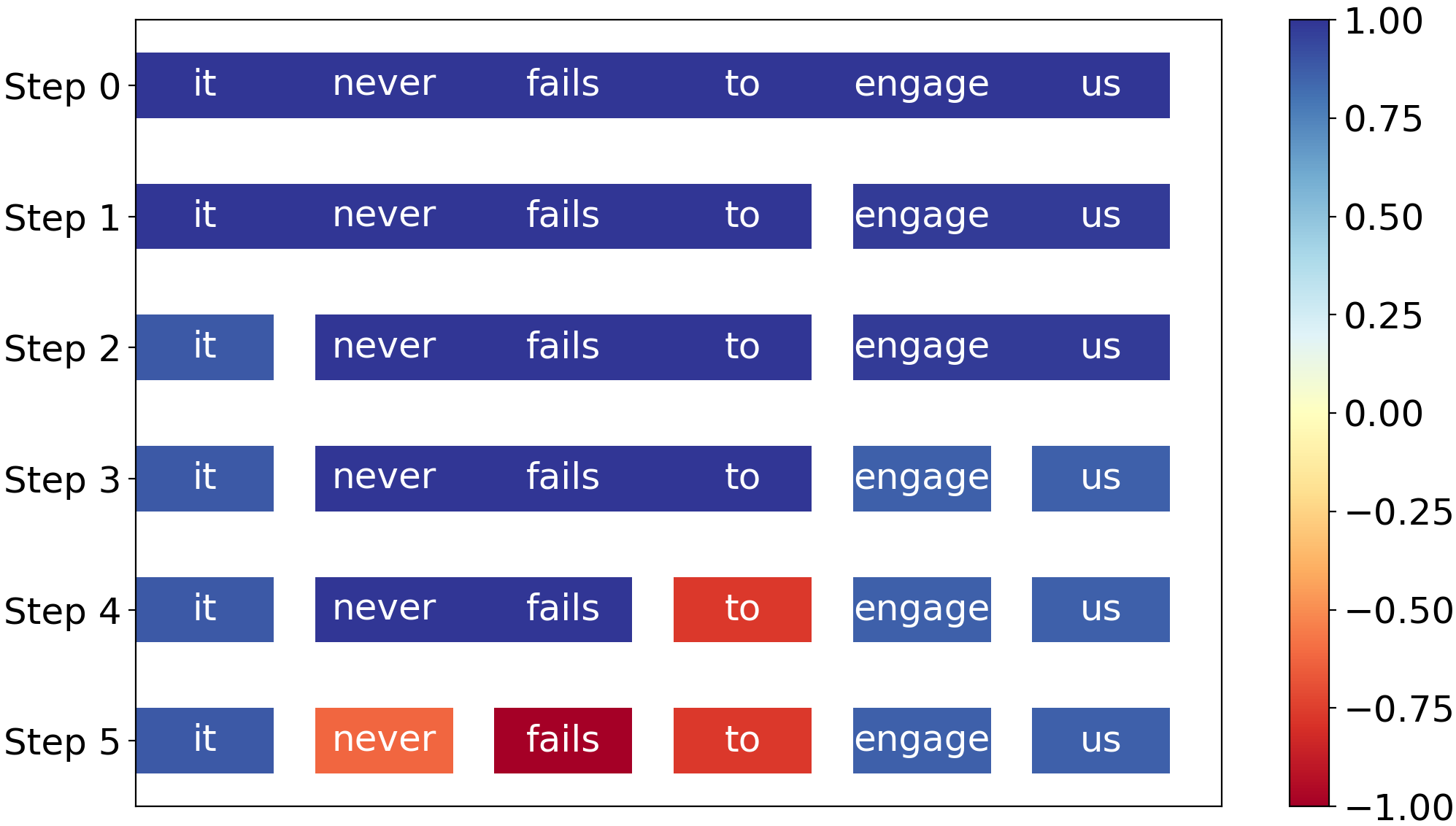}}
  \caption{\label{fig:inter_bert_2} \ourmethod for BERT on a positive movie review from the SST dataset. BERT makes the correct prediction because it captures the interaction between \texttt{never} and \texttt{fails}.}      
\end{figure}

\begin{figure}[H]        
  \center{\includegraphics[width=7.5cm]  {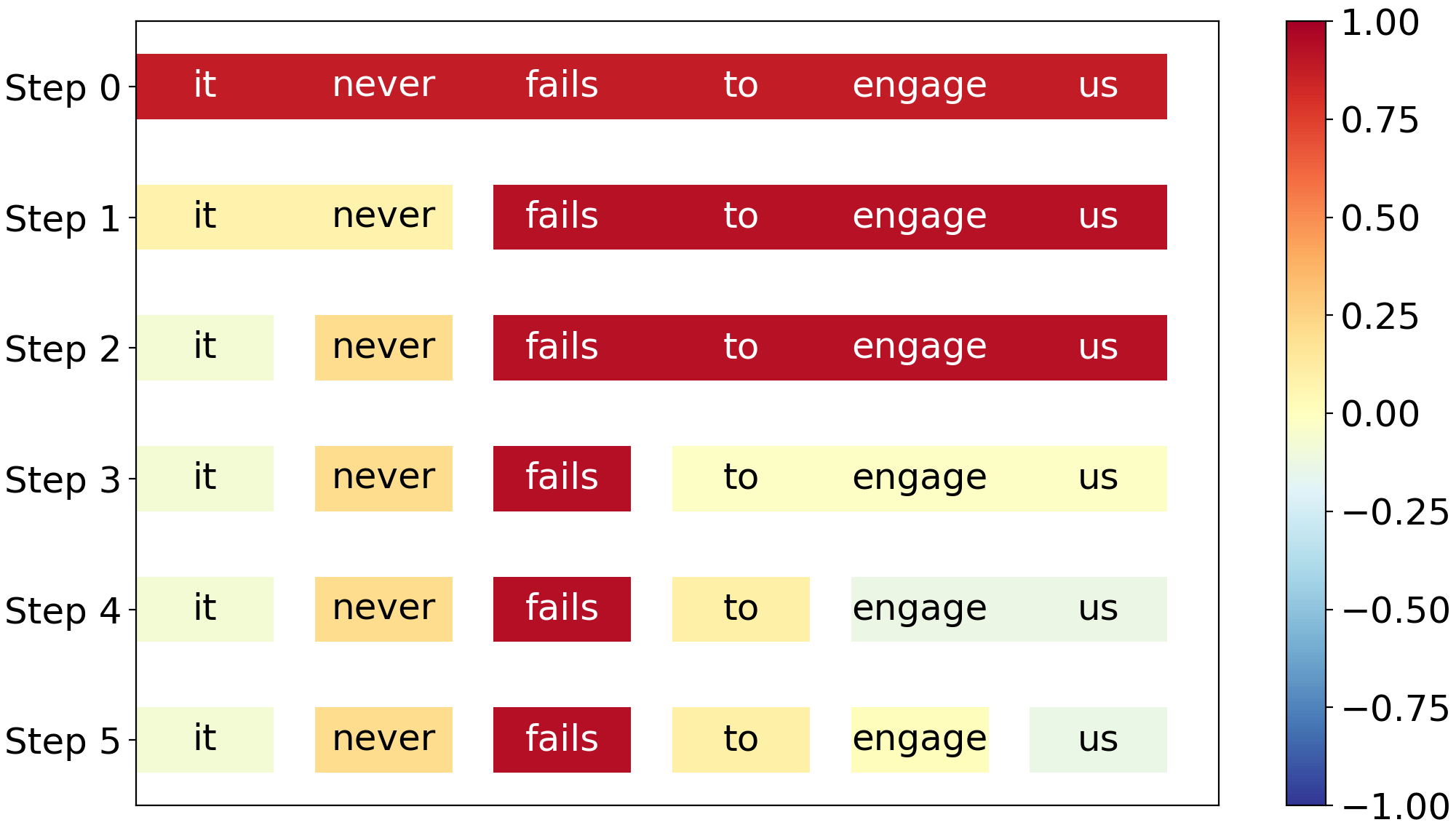}}
  \caption{\label{fig:inter_lstm_2} \ourmethod for LSTM on a positive movie review from the SST dataset. LSTM makes the wrong prediction because it misses the interaction between \texttt{never} and \texttt{fails}.}      
\end{figure}

\begin{figure}[H]        
  \center{\includegraphics[width=7.5cm]  {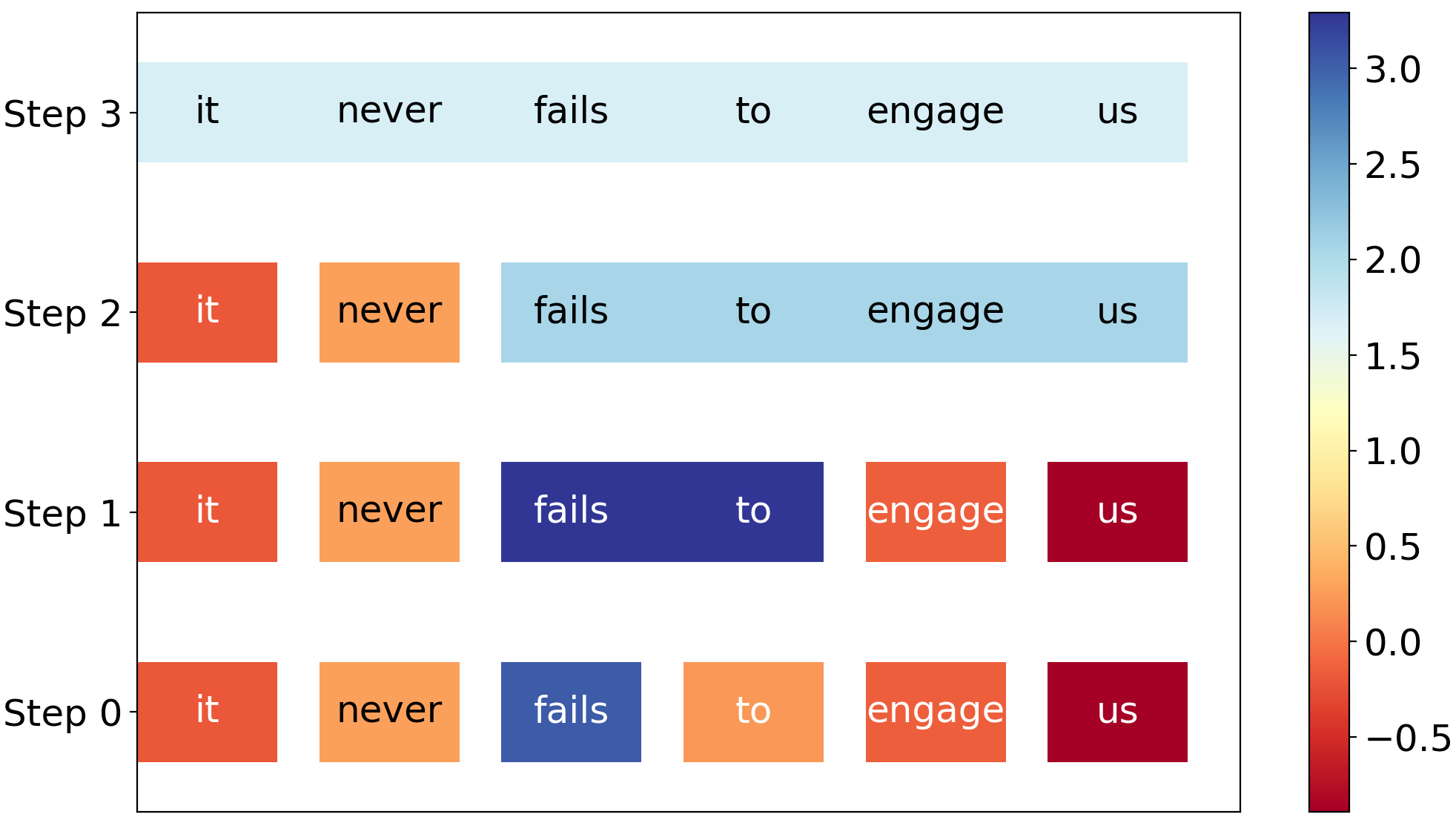}}
  \caption{\label{fig:ACD_lstm_2} ACD for LSTM on a positive movie review from the SST dataset, on which LSTM makes wrong prediction.}      
\end{figure}

\begin{figure}[H]        
  \center{\includegraphics[width=7.5cm]  {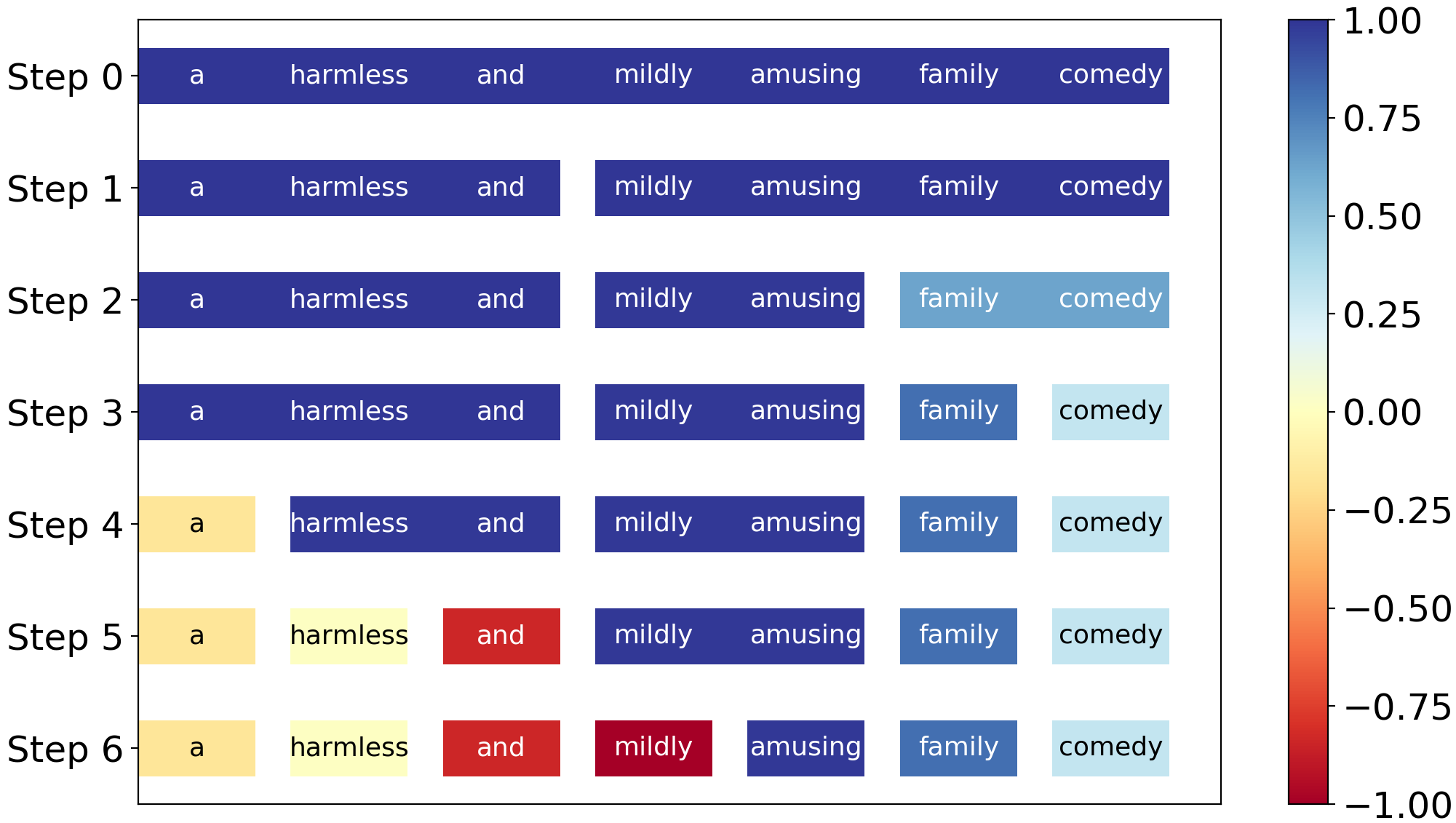}}
  \caption{\label{fig:inter_bert_1} \ourmethod for BERT on a positive movie review from the SST dataset, on which BERT makes correct prediction.}      
\end{figure}

\begin{figure}[H]        
  \center{\includegraphics[width=7.5cm]  {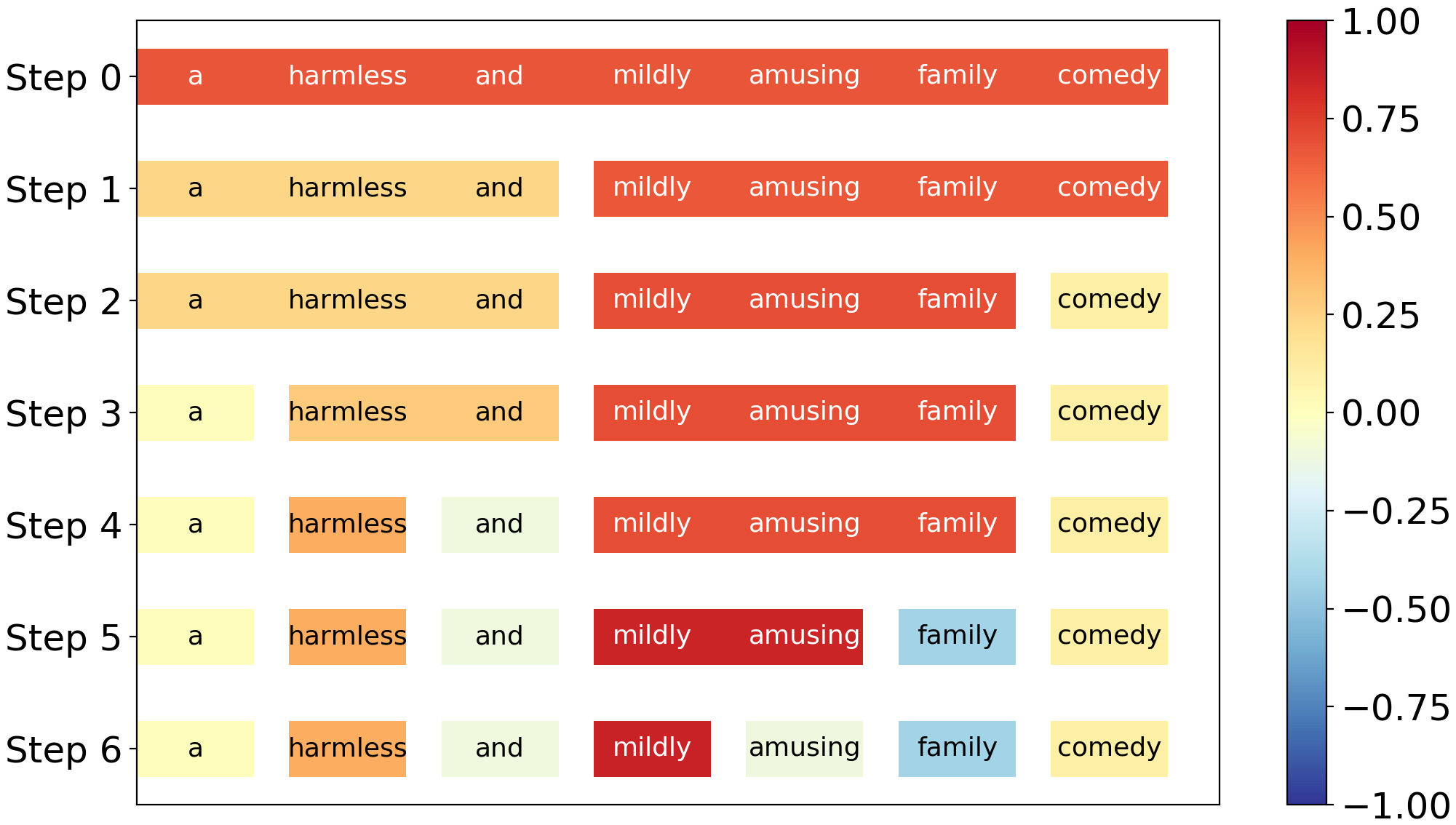}}
  \caption{\label{fig:inter_lstm_1} \ourmethod for LSTM on a positive movie review from the SST dataset, on which LSTM makes wrong prediction.}      
\end{figure}

\begin{figure}[H]        
  \center{\includegraphics[width=7.5cm]  {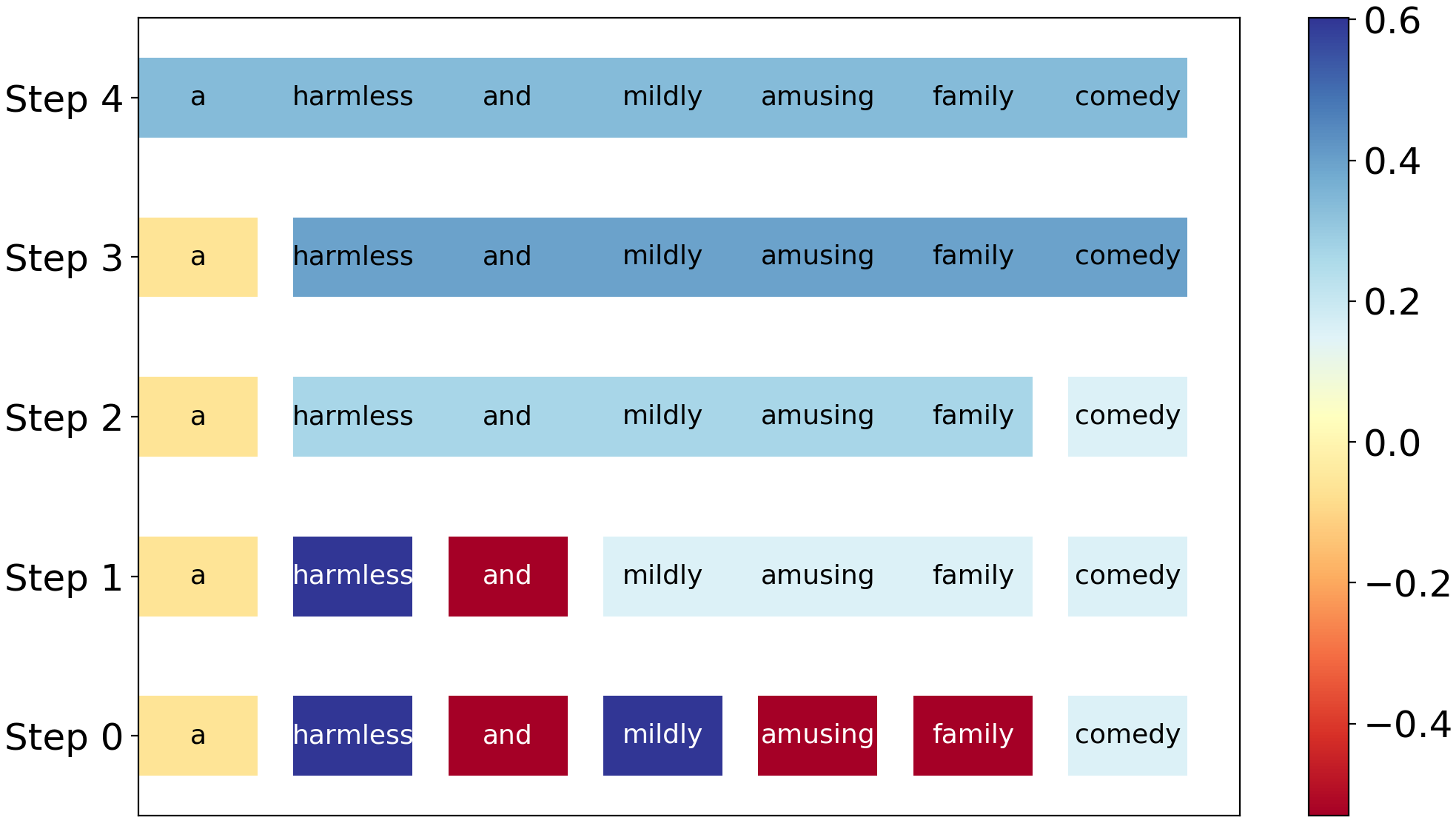}}
  \caption{\label{fig:ACD_lstm_1} ACD for LSTM on a positive movie review from the SST dataset, on which LSTM makes wrong prediction.}      
\end{figure}

\section{Human Evaluation Interface}
\label{sec:interface}
\begin{figure*}[th!]        
  \center{\includegraphics[width=10cm]  {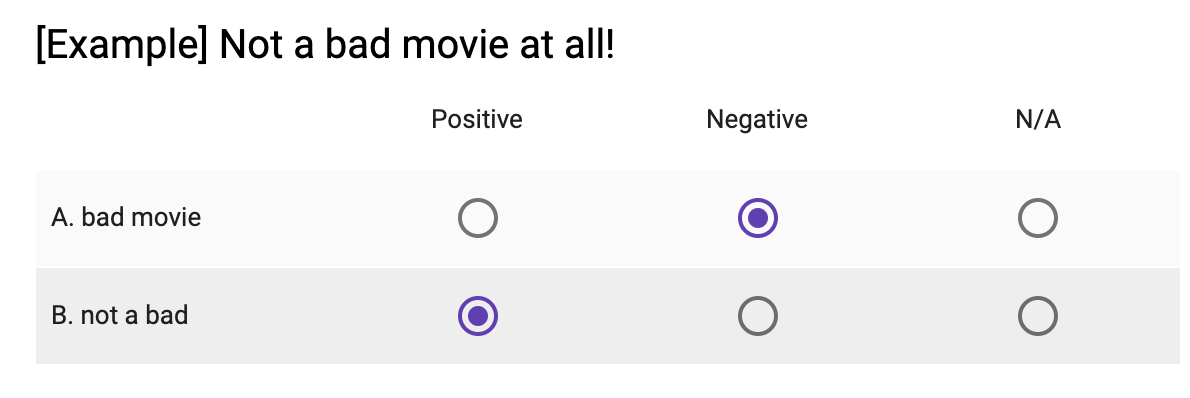}}
  \caption{\label{fig:interface} Interfaces of Amazon Mechanical Turk where annotators are asked to guess the model's prediction based on different explanations.}      
\end{figure*}


\end{document}